\PassOptionsToPackage{table}{xcolor}
\documentclass[acmsmall,screen]{acmart}

\usepackage{float} % in preamble
\usepackage{siunitx}

\AtBeginDocument{%
  }

\setcopyright{cc}
\setcctype{by}
\acmJournal{TELO}
\acmYear{2026}
\acmMonth{1} \acmDOI{10.1145/3845808}
\received{14 November 2025}
\received[revised]{30 April 2026}
\received[accepted]{12 August 2026}

\begin{document}

\title[Transformer Semantic Genetic Programming for $d$-dimensional Symbolic Regression Problems]{Transformer Semantic Genetic Programming\\ for $d$-dimensional Symbolic Regression Problems}

\author{Philipp Anthes}
\email{anthes@uni-mainz.de}
\orcid{0009-0009-8383-6551}
\affiliation{%
  \institution{Johannes Gutenberg University Mainz}
  \city{Mainz}
  \country{Germany}
}

\author{Dominik Sobania}
\email{dominik.sobania@uni-due.de}
\orcid{0000-0001-8873-7143}
\affiliation{%
  \institution{paluno, University of Duisburg-Essen}
  \city{Essen}
  \country{Germany}
}
\author{Franz Rothlauf}
\email{rothlauf@uni-mainz.de}
\orcid{0000-0003-3376-427X}
\affiliation{%
  \institution{Johannes Gutenberg University Mainz}
  \city{Mainz}
  \country{Germany}
}

\renewcommand{\shortauthors}{Anthes et al.}
\acmArticleType{Research}

\begin{abstract}
Transformer Semantic Genetic Programming (TSGP) is a semantic search approach that uses a pre-trained transformer model as a variation operator to generate offspring programs with high semantic similarity to a given parent. Unlike other semantic GP approaches that rely on fixed syntactic transformations, TSGP aims to learn diverse structural variations that lead to solutions with similar semantics. We find that a single transformer model trained on millions of programs is able to generalize across symbolic regression problems of varying dimensions. Evaluated on 24 real-world and synthetic datasets, TSGP significantly outperforms standard GP, SLIM\_GSGP, Deep Symbolic Regression, and Denoising Autoencoder GP, achieving an average rank of 1.58 across all benchmarks. Moreover, TSGP produces more compact solutions than SLIM\_GSGP, despite its higher accuracy. In addition, the target semantic distance is able to effectively adjust the step size in the semantic space: small values enable consistent improvement in fitness but often lead to larger programs, while larger values promote faster convergence and compactness. Thus, the target semantic distance provides an effective mechanism for balancing exploration and exploitation.
\end{abstract}




\maketitle
\newpage
\section{Introduction}\label{Introduction}
Symbolic regression (SR) aims to identify programs that accurately capture the underlying relationships in data. In SR, programs are represented as symbolic expressions constructed from operators (e.g. addition, multiplication), variables (e.g. $x_1$,$x_2$,$x_3$), and constants. The resulting syntactical composition embodies a certain program behavior that determines the way predictions (output values) are generated from the available data (input values) and, consequently, how the program fits to the underlying data. 

Genetic programming (GP) has developed several approaches to search for high-quality programs. For example, standard GP (stdGP) uses syntactic variation operators completely ignoring the impact of syntactic modifications on program behavior. As a result, even minor syntactic changes can produce offspring with significantly different behavior, which often disrupts promising programs, leading to low search efficiency \citep{uy2009semantics,moraglio_geometric_2012,mcphee2008semantic}. To address this limitation, semantic-aware approaches instead incorporate program behavior (semantics) to guide the variation process. One of the best-performing semantic approaches is Geometric Semantic Genetic Programming (GSGP), which uses Geometric Semantic Operators (GSOs) to generate offspring that are semantically similar to their parents \citep{moraglio_geometric_2012}. GSOs have been shown to be very effective variation operators, as they induce a smooth, unimodal fitness landscape for the underlying problem \citep{vanneschi2016introduction}. 

However, GSOs operate by performing linear combinations of program structures, independent of the syntactic form of the parent. This reliance on fixed, predefined transformation rules limits their ability to express semantic similarity through diverse and adaptive syntactic variations. Consequently, GSO-based approaches tend to produce overly complex and bloated programs \citep{anthes2025transformer,vanneschi_slim_gsgp_2024,martins2018solving}. GSOs do not take advantage of the fact that the same semantics can be achieved by many different syntactic forms and that numerous structurally efficient variations yield similar program behavior. An effective search strategy should operate directly in semantic space, independently of its syntactic representation.

A recently introduced approach, Transformer Semantic Genetic Programming (TSGP), employs a transformer-based variation operator trained on millions of semantic variations of programs to generate semantically similar offspring \citep{anthes2025transformer}. In an offline \textit{Model Building} step, a transformer model is trained to learn which syntactic modifications lead to similar semantic behavior. During training, pairs of programs with similar semantics are formed: the model takes one program as input and aims to synthesize the other as the target output, conditioned on the semantic distance ($\mathrm{SD}$) between them. After training, the transformer acts as a zero-shot semantic variation operator: given a parent program and a target semantic distance ($\mathrm{SD}_t$), it generates a new offspring program that is semantically similar to the parent, where $\mathrm{SD}_t$ influences the degree of semantic similarity. Transformers are particularly well suited for this difficult task, having demonstrated their ability in numerous domains to capture contextual relationships in sequences and generate results that correspond to the desired behaviors, regardless of syntactic structure \citep{vaswani2017attention,brown2020language,briesch2023large}.

Previous work introduced TSGP for fixed-dimension problems ($d=4$) and showed its superiority in quality and compactness \citep{anthes2025transformer}. This article extends that work in two key directions:
\begin{enumerate}
    \item We show that a single transformer model can be trained to handle SR problems of varying dimensions $d$. By conditioning the model on the dimension $d$ of the target problem, we ensure that valid expressions are generated, using at most $d$ input variables. This enables evaluation across a comprehensive benchmark set of 24 real-world and synthetic SR problems of various dimensions.
    \item We study how the target semantic distance $\mathrm{SD}_t$ influences the search process. We analyze its impact on resulting convergence speed and program size and demonstrate that adjusting $\mathrm{SD}_t$ effectively modulates the step size in the semantic space, thereby offering a mechanism to regulate the trade-off between exploration and exploitation. 
\end{enumerate}
In a comprehensive experimental study, we compare the performance of TSGP against stdGP, SLIM\_GSGP, and two model-based search approaches: Deep Symbolic Regression (DSR) \citep{petersen_deep_2021} and Denoising Autoencoder GP (DAE-GP) \citep{wittenberg_dae-gp_2020}. The results show that TSGP finds solutions of significantly higher prediction quality than the benchmark approaches, achieving an average rank of 1.58 across all datasets. Furthermore, TSGP significantly outperforms SLIM\_GSGP, a GSGP variant with a deflation operator, in solution compactness. Our study of the target semantic distance $\mathrm{SD}_t$ reveals that it effectively influences the step size in the semantic space. A small $\mathrm{SD}_t$ results in small semantic steps, enabling consistent fitness improvements in each generation. However, as expected, this often results in slower convergence and limited exploration, thus producing larger programs. In contrast, a larger $\mathrm{SD}_t$ encourages broader exploration, leading to smaller and more compact programs with faster initial convergence. Yet, this comes at the cost of reduced exploitation, which can have a negative impact on the quality of the final solution.

Sect.~\ref{sec:RelatedWork} reviews related work in semantic GP, model-based search approaches, and transformer-based one-shot estimators. Sect.~\ref{sec:TSGPApproach} presents the TSGP approach, including \textit{Model Building}~\ref{sec:Model_Building} and \textit{Model Inference} ~\ref{sec:Model_Inference}. Sect.~\ref{sec:Experimental_Settings} describes the experimental setup used to evaluate the approaches. Sect.~\ref{sec:Experiments} reports the results, focusing on solution quality and program size, and further analyzes how the semantic step size is affected by different values of the target semantic distance $\mathrm{SD}_t$. Sect.~\ref{sec:Limitations_Future_Work} suggests directions for future research, and Sect.~\ref{sec:Conclusions} summarizes the key findings.

\section{Related Work}\label{sec:RelatedWork}
\paragraph{Semantic GP} Semantic-aware approaches in GP aim to control the semantic similarity between offspring and their parent programs \citep{vanneschi2014survey}. We focus specifically on methods that, like TSGP, generate new offspring from a single parent program.
Early approaches indirectly influenced semantics during variation by rejecting offspring based on semantic criteria. For example, Semantically Driven Mutation prevents the offspring from being semantically identical to the parent \citep{beadle2009semantically}, while Semantic Similarity-based Mutation discards offspring that deviate significantly from the parent’s semantics \citep{uy2009semantics}. Both approaches increase the performance of GP by consistently producing promising candidate solutions.

The first semantic operators that directly generate offspring of controlled and weighted semantic similarity were introduced with GSGP \citep{moraglio_geometric_2012}. Geometric Semantic Mutation creates a new program by appending a program structure to the parent considering a mutation step size. The appended structure consists of two randomly generated programs that are subtracted from each other, ensuring only minor semantic modifications to the parent. Due to the efficient implementation suggested by \citep{vanneschi_new_2013}, GSGP could be successfully applied to regression tasks in domains such as pharmacokinetics, concrete strength prediction, and financial data analysis \citep{Mcdermott_GSGP_Financial_Data, castelli2013prediction, vanneschi_geometric_2014}.

However, repeated addition of program structures leads to exponential growth in program size during the search. As a result, approaches have been proposed to mitigate this growth:
One such approach is GSGP-Red, which simplifies programs by merging repetitive structures and coefficients that occur after applying GSOs \citep{martins2018solving}. In contrast, SLIM\_GSGP argues that semantic similarity can be achieved not only by adding randomly generated structures with minimal semantic impact but also by removing them in later stages of the search. Therefore, SLIM\_GSGP introduced a new geometric semantic deflation operator that systematically reduces the size of the offspring \citep{vanneschi_slim_gsgp_2024}.
\paragraph{Neural Network-Guided Search.} In recent years, neural networks have been proposed as search operators for symbolic regression tasks. By using feedback from previous search steps, these search operators generate new programs auto-regressively, token by token.

DSR uses reinforcement learning to train a recurrent neural network (RNN) during the search process \citep{petersen_deep_2021}. The RNN acts as a policy network that sequentially samples symbolic expressions according to its learned policy. Each completed expression is evaluated on the target problem, and the resulting error is used to optimize the model parameters of the RNN. Later variants improve performance by adding large-scale pre-training or hybrid search strategies, such as neural-guided GP at decoding time \citep{Landajuela_2022}.

In contrast, DAE-GP is an estimation-of-distribution algorithm that uses a Long Short-Term Memory (LSTM) denoising autoencoder to model the distribution of high-quality programs in the latent space of the LSTM \citep{wittenberg_dae-gp_2020}. Once the LSTM has learned the distribution of the selected population, the LSTM is used to create a new offspring population: parental programs are slightly perturbed and passed through the network, generating new programs with similar syntactic properties. For SR problems, DAE-GP has demonstrated its ability to produce programs of similar quality compared to standard syntactic operators, while significantly reducing the complexity and size of the resulting structures \citep{wittenberg_small_2023}.

Both DSR and DAE-GP operate in an online manner, where neural networks are trained and optimized during the search. In contrast, TSGP’s model-based search operates offline and does not require training during the search, enabling faster and more efficient generation of new programs.
\paragraph{One-shot Transformers.} 
Recent transformer-based approaches to symbolic regression operate in a one-shot manner. 
Given input-output data, such models generate candidate solutions in a single forward pass without an iterative search or refinement of programs \citep{biggio_neural_2021,kamienny2022end}. The models are trained on large synthetic datasets consisting of numerical inputs, symbolic expressions, and their corresponding outputs. During training, the transformer learns to map input-output pairs to symbolic expressions. At inference time, it produces a solution via a static direct mapping, rather than exploring the solution space \citep{biggio_neural_2021,kamienny2022end}.

However, recent work demonstrates that one-shot models have great difficulty to generalize beyond their pre-training distribution and are not competitive with search-based baselines \citep{voigt2025analyzing}. Unlike search-based methods (such as stdGP, SLIM\_GSGP or TSGP), which iteratively explore the space of possible programs, one-shot transformers rely entirely on learned input-output-expression mappings. Furthermore, while many of these models are publicly available, the used training datasets are not published, making it impossible to assess whether test problems were inadvertently seen during training.
\begin{figure}[t]
    \centering
    \includegraphics[width=1\linewidth]{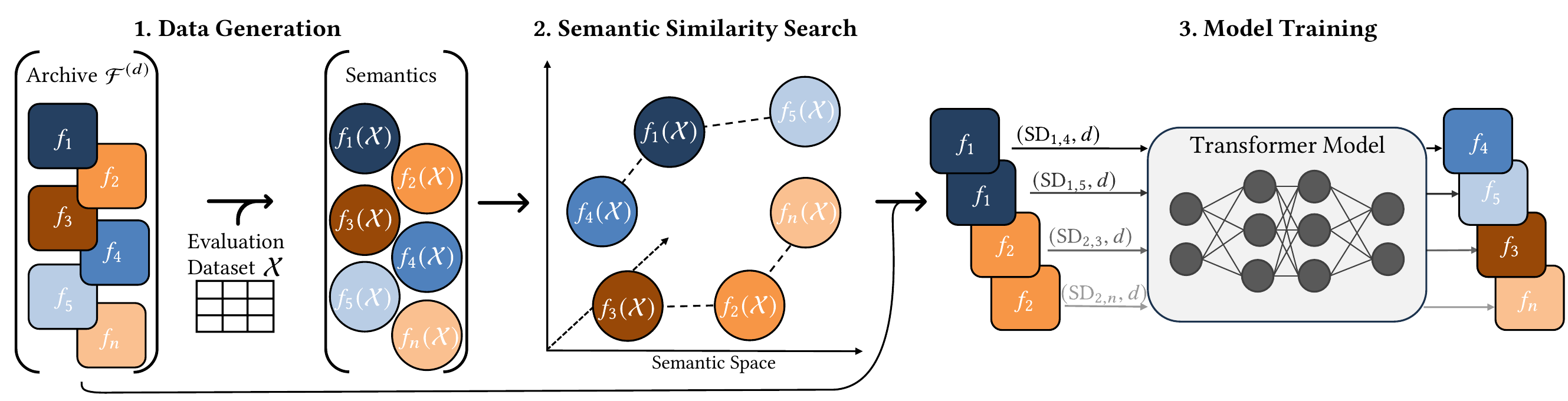}
\caption{\textit{Model Building} of TSGP: (1) Diverse functions are generated and their semantics are approximated; (2) Semantically similar pairs are identified through a $k$-NN search in the semantic space; (3) These pairs are used as input-output examples to train a transformer model, conditioned on their semantic distance $\mathrm{SD}$ and the problem dimensionality $d$.} \label{fig:TSGP_Model}
\end{figure}

\section{The TSGP Approach}\label{sec:TSGPApproach}
TSGP is a semantic search approach that uses a pre-trained transformer model as a variation operator to generate semantically similar offspring. In a single, offline \textit{Model Building} step, the model captures structural patterns between symbolic expressions (referred to as mathematical functions) that generate high semantic similarity. During \textit{Model Inference}, the transformer model can transfer these learned patterns to mathematical functions for unknown datasets without being fine-tuned to the task.

\subsection{Model Building} \label{sec:Model_Building}
\textit{Model Building} in TSGP consists of three steps:
First, a diverse set of mathematical functions is generated. The generated functions should well cover the possible search space. Second, semantically similar function pairs are identified and grouped to input-output training examples. Third, a transformer model is trained in a sequence-to-sequence task to generate a semantically similar output sequence based on the given input sequence. Each step is illustrated in Figure~\ref{fig:TSGP_Model} and is described below.

\paragraph{Data Generation.} 
In the first step of \textit{Model Building}, we generate the training data used to pre-train the transformer model. Unlike \citep{anthes2025transformer}, which trained a model for fixed-dimension problems, we build a single transformer model capable of generalizing across SR problems of varying dimensionality. Thus, instead of relying on a single archive of fixed dimension, we now create multiple archives \(\mathcal{F}^{(d)}= \{f_1, f_2, \dots, f_{n_d}\}\), where each archive consists of $n_d$ unique symbolic functions valid for $d$-dimensional SR problems. The functions in each archive $\mathcal{F}^{(d)}$ are obtained by applying stdGP to a diverse set of $d$-dimensional, synthetically generated SR problems. During each GP run, numerous functions valid for $d$-dimensional problems are generated, which are all collected in $\mathcal{F}^{(d)}$. The underlying SR problems are constructed from linear regression models with added Gaussian noise and inputs are sampled from a standard normal distribution. Additionally, we convert the functions in $\mathcal{F}^{(d)}$ into a uniform form using SymPy \citep{Sympy}. This step reduces the number of syntactically redundant expressions and thus provides the transformer model with a clearer training signal regarding which syntactic changes actually affect the semantics, as shown in Appendix~\ref{appendix:Ablation_Studies}. Note that on average this conversion does not result in a reduction in function size.

In the literature, semantics is commonly defined as the behavior a function exhibits when evaluated on a given set of inputs \citep{krawiec2009approximating,uy2011semantically}. To ensure consistency between the semantics measured during \textit{Model Building} and those during evolutionary search, we define a standardized input space (where all features are standardized to zero mean and unit variance) that matches the input distributions used in later experiments (Section~\ref{sec:Experimental_Settings}).
Specifically, we approximate the semantics of each function \(f \in \mathcal{F}^{(d)}\) by evaluating it on a fixed, randomly sampled dataset \(\mathcal{X} = [\mathbf{x}_1, \mathbf{x}_2, \dots, \mathbf{x}_m]\), where each $\mathbf{x}_i \in \mathbb{R}^d$ is drawn independently from a standard normal distribution across all $d$ input dimensions. With $m$ sufficiently large, $\mathcal{X}$ densely samples the standardized input space, ensuring that the resulting output vector \(f(\mathcal{X}) = (f(\mathbf{x}_1), f(\mathbf{x}_2),\dots, f(\mathbf{x}_m))\) forms a semantic vector $\mathbf{s}(f)\in\mathbb{R}^m$, which serves as a comprehensive numerical approximation of $f$'s true behavior under standardized conditions. 

\paragraph{Semantic Similarity Search.}
Using the computed semantic vectors, semantically similar functions are identified through a semantic similarity search. Two functions $f_i, f_j\in \mathcal{F}^{(d)}$ are considered semantically similar if the distance between their semantic vectors $\mathbf{s}(f_i)$ and $\mathbf{s}(f_j)$ is small \citep{uy2011semantically}. Since semantics is represented using vectors in $\mathbb{R}^m$, similarity can be quantified using a distance measure \citep{moraglio_geometric_2012}. We use the Euclidean distance and define the semantic distance $\mathrm{SD}$ between two semantic vectors $f_i$ and $f_j$ as $\mathrm{SD}(f_i, f_j) = \|\mathbf{s}(f_i) - \mathbf{s}(f_j)\|_2$.
To identify similar functions, we perform a $k$-nearest neighbor ($k$-NN) search on the semantics for each archive $\mathcal{F}^{(d)}$. For each function \(f_i \in \mathcal{F}^{(d)}\), the $k$-NN search returns a set of $k$ neighbors $N_k(f_i)$ with minimal semantic distance:
\begin{equation}
N_k(f_i) = \underset{\{\mathcal{F}' \subset \mathcal{F}^{(d)}, |\mathcal{F}'| = k\}}{\operatorname{argmin}} \sum_{f \in \mathcal{F}'} \mathrm{SD}(f_i, f).
\end{equation}
From these sets of neighbors, we generate $k$ input-output training pairs $(f_i,f_o)$ per function, where $f_i$ is the input, and each \(f_o\in N_k(f_i)\) serves as a semantically similar target output.

The similarity search is performed separately for each archive $\mathcal{F}^{(d)}$. This yields $ |\mathcal{F}^{(d)}| \times k $ input-output training pairs per archive. Given $D$ archives of comparable size, the total number of generated training pairs is roughly $D \times |\mathcal{F}^{(d)}| \times k $. 

\paragraph{Model Training.} 
In the third step, an encoder-decoder transformer model is trained on all input-output pairs $(f_i,f_o)$ collected from the semantic similarity searches across all archives $\mathcal{F}^{(d)}$. The model uses the architecture proposed by \citep{vaswani2017attention} and is trained from scratch exclusively for symbolic expression data using random model parameters \(\theta\) for initialization.

The main training objective is to generate the semantically similar output function $f_o$ given the input function $f_i$. To achieve this, the input function is first encoded into the transformer's latent representation. The output function is then generated step by step in an auto-regressive manner: at each step, the model predicts the next token based on the latent representation of the input and the tokens already generated from the current sequence. The model parameters \(\theta\) are updated in batches by minimizing the cross-entropy loss between the predicted token and the target token of the output sequence $f_o$. Upon convergence, the optimized parameters \(\theta^*\) are stored for inference.

During training, two additional inputs are provided alongside $f_i$ to guide the semantic similarity and the dimensional validity of the output. The first is the semantic distance $\mathrm{SD}(f_i, f_j)$ between the input-output pair, which allows the model to quantitatively learn semantic similarity. In inference, conditioning the model on a target semantic distance $\mathrm{SD}_t$ influences how closely the generated offspring matches the parent's behavior. The second input is the dimensionality $d$ of the archive $\mathcal{F}^{(d)}$ from which the pair was drawn. Since the model only receives function pairs of dimensionality less than or equal to $d$, it learns to generate expressions that depend on not more than $d$ variables. In inference, setting $d$ to the dimensionality of the problem conditions the model to produce outputs that are valid for the input space.

The result of \textit{Model Building} is a single transformer model that can generate expressions with adjustable semantic similarity to a given input without further fine-tuning, across problems of varying dimensions.

\subsection{Model Inference} \label{sec:Model_Inference}
After \textit{Model Building}, the pre-trained transformer model with parameters \(\theta^*\) is used during the evolutionary search as a semantic variation operator, replacing the standard GP variation operators. 

During variation, the model independently generates for each parent function $f_i$ a new offspring $f_o$, conditioned on the target semantic distance $\mathrm{SD}_t$ and the problem dimensionality $d$. The parameter $\mathrm{SD}_t$ influences how closely the behavior of the offspring matches that of the parent, thus allowing the step size in the semantic space to be regulated. Adding the dimensionality of the problem $d$ conditions the model to use only variables $x_1$ through $x_d$ during inference, resulting in compatible expressions for the $d$-dimensional input of the target problem. 

Function generation proceeds auto-regressively: the model simultaneously predicts the next token for every function in the batch, conditioned on the respective parent function and the context of the tokens generated so far. Inference stops when either an end-of-sequence (EOS) token is created for every function in the batch, or the maximum sequence length is reached. To ensure syntactic validity of the generated output, we employ syntax control as proposed by \citep{wittenberg2022using}, which enforces syntactic correctness during auto-regressive generation. At every decoding step, token probabilities are modified to assign zero probability to syntactically invalid options, such as operators with incorrect arity.

\section{Experimental Settings}\label{sec:Experimental_Settings}
\paragraph{Datasets.}
We compare the performance of TSGP with stdGP, SLIM\_GSGP, DSR, and DAE-GP on 24 SR problems taken from Penn Machine Learning Benchmarks \citep{romano2021pmlb}, a widely used benchmark suite in SR research, which is also included in large-scale studies such as SRBench \citep{la2021contemporary}. For our experiments, we select all available real-world datasets with a dimensionality ranging from 2 to 5 dimensions where a sufficient number of samples ($>100$) are available. We complement these 12 real-world datasets with 12 synthetic datasets, including the Pollen dataset used in \citep{anthes2025transformer} and randomly sampled Feynman datasets spanning various dimensionalities. On the Feynman datasets, solution quality is evaluated using a randomly drawn subsample of 10,000 records (without replacement) from the full dataset for each independent run, to ensure computational feasibility and reliable evaluation. The datasets are listed in Table~\ref{tab:datasets}.

Before training and evaluation, the feature and target variables in the datasets are standardized to have zero mean and unit variance. This is in accordance with the standardized input space used during TSGP’s \textit{Model Building} step, ensuring the transferability of the pre-trained model to datasets of different value ranges. Standardization also improves solution quality and reduces bloat in GP-based methods by making the search process invariant to the scale of input features and constants \citep{owen2018feature,dick2020feature}. 

\begin{table}
\centering
\caption{The benchmark datasets consist of 12 real-world and 12 synthetic problems with dimensionality ranging from 2 to 5 and sample sizes from 100 to 10,000.}
\label{tab:datasets}
\begin{tabular}{p{0.3\textwidth}|p{0.2\textwidth}|p{0.2\textwidth}|p{0.2\textwidth}}
\hline
\rowcolor{gray!20}
\textbf{Dataset} & \textbf{\# Samples} & \textbf{Dimensionality} & \textbf{Type}\\ 
\hline
Analcatdata Apnea1 & 475 & 3 & Real-world\\
\rowcolor{gray!10}
Analcatdata Apnea2 & 475 & 3 & Real-world\\
Analcatdata Neavote & 100 & 2 & Real-world\\
\rowcolor{gray!10}
Chscase Geyser1 & 222 & 2 & Real-world\\
Cloud & 108 & 5 & Real-world\\
\rowcolor{gray!10}
ERA & 1,000 & 4 & Real-world\\
ESL & 488 & 4 & Real-world\\
\rowcolor{gray!10}
Feynman I 18 4 & 10,000 & 4 & Synthetic \\
Feynman I 24 6 & 10,000 & 4 & Synthetic \\
\rowcolor{gray!10}
Feynman I 25 13 & 10,000 & 2 & Synthetic \\
Feynman I 29 4 & 10,000 & 2 & Synthetic \\
\rowcolor{gray!10}
Feynman I 43 43 & 10,000 & 4 & Synthetic \\
Feynman II 34 2 & 10,000 & 3 & Synthetic \\
\rowcolor{gray!10}
Feynman II 38 3 & 10,000 & 4 & Synthetic \\
Feynman II 4 23 & 10,000 & 3 & Synthetic \\
\rowcolor{gray!10}
Feynman III 14 14 & 10,000 & 5 & Synthetic \\
Feynman Test 3 & 10,000 & 4 & Synthetic \\
\rowcolor{gray!10}
Feynman Test 4 & 10,000 & 5 & Synthetic \\
Galaxy & 323 & 4 & Real-world\\
\rowcolor{gray!10}
LEV & 1,000 & 4 & Real-world\\
Pollen & 3,848 & 4 & Synthetic \\
\rowcolor{gray!10}
Rabe & 120 & 2 & Real-world\\
Vinnie & 380 & 2 & Real-world\\
\rowcolor{gray!10}
Visualizing Environmental & 111 & 3 & Real-world\\
\hline
\end{tabular}
\end{table}
\begin{table}
\centering
\caption{Configuration parameters and values used for the evaluated methods.}
\label{tab:gp_parameters}
\begin{tabular}{p{0.3\textwidth}|p{0.4\textwidth}} 
 \hline
 \rowcolor{gray!20}
 \textbf{Parameter} & \textbf{Value} \\ 
 \hline
 \rowcolor{white}
 Initialization & Ramped Half-and-Half\\
 \rowcolor{gray!10}
Primitive Set & $\{V, \text{ERC}, +, -, \times, \%, \text{pow}\}$\\
 \rowcolor{white}
 ERC Range & $[-0.5, 0.5]$, step size 0.1\\ 
 \rowcolor{gray!10}
 Population Size & 100 \\ 
\rowcolor{white}
 Generations & 100 \\ 
\rowcolor{gray!10}
 Selection & Tournament selection of size 5 \\ 
 \rowcolor{white}
 Evaluation Metric & RMSE\\ 
\rowcolor{gray!10}
 Runs & 30 \\ 
 \hline
\end{tabular}
\end{table}
\paragraph{Search Settings.}
For TSGP, stdGP and DAE-GP, we use the evolutionary computation framework DEAP \citep{fortin2012deap}, with neural networks being implemented with Keras \citep{chollet2015keras}. For SLIM\_GSGP and DSR, we use publicly available Python implementations \citep{vanneschi_slim_gsgp_2024, petersen_deep_2021}. GP parameters that are common in all methods are described in Table~\ref{tab:gp_parameters}.

Initial populations are generated using Ramped Half-and-Half (RHH). For TSGP, stdGP and DAE-GP, which are based on the DEAP framework, we choose an initialization depth between 2 and 5. For SLIM\_GSGP, we rely on the default initialization values specified by the framework. We set the maximum allowed tree depth during a run to 17 ~\citep{koza1992programming}. 

The terminal set includes all input variables $V = \{v_1, v_2, \dots v_d\}$ that correspond to the dimensionality $d$ of the dataset and ephemeral random constants (ERCs) sampled from $[-0.5, 0.5]$ in steps of 0.1. The function set consists of addition, subtraction, multiplication, protected division (returning 1 on division by zero) and pow.
We include pow because SymPy represents division as exponentiation \citep{Sympy}, which leads to having pows in the training data of TSGP. Although the pow operator is not available in the SLIM\_GSGP or DSR frameworks, its inclusion in stdGP did not positively affect performance compared to previous results \citep{anthes2025transformer}.

The population size is set to 100 and the search is conducted for 100 generations. Parents are selected via tournament selection with a tournament size of 5 and no elitism is applied. Each dataset is divided into training and test sets in a 75\% to 25\% ratio. We use Root Mean Squared Error (RMSE) as the evaluation metric, which quantifies the Euclidean distance between the function's semantics \(f(\mathcal{X}) \) and the target semantics. The best program (i.e., the one with the lowest training RMSE) is extracted as the final solution and evaluated on the test set to measure its prediction quality. 

\paragraph{TSGP-Setup.}
To create training data for the transformer model of TSGP, we apply stdGP to 50 synthetic regression problems for each dimensionality $d \in \{2,3,4,5\}$. Each stdGP run uses a population of 2,000 and continues until 150,000 unique functions are collected. All other parameters match those specified in Table~\ref{tab:gp_parameters}. To balance accuracy and simplicity and reduce bloat during evolution, we employ double tournament selection \citep{Sean2002FightingBloat}. 

Pairs of programs with high semantic similarity are obtained using a $k$-nearest neighborhood search ($k=3$) based on the Euclidean distance between semantic vectors, each computed on a fixed, standardized dataset of 500 input points. To enable an efficient similarity search, we use the FAISS library \citep{douze2024faisslibrary}, which reduces computational costs by clustering semantic vectors and computing distances only within each cluster. An input-output pair $(f_i,f_o)$ is created for each function $f_i$ and its neighbor $f_o$ if $0 <\mathrm{SD}(f_i, f_o) < 100$. This excludes semantically identical functions, which would not result in an improvement in fitness during variation, as well as outlier pairs with extreme semantic distances that lie far outside of the actual distribution (as shown in Figure \ref{fig:Distances_Trainingdata}). Additionally, functions should not exceed a maximum length of 100 tokens to ensure computational feasibility during inference. For each dimension $d \in \{2,3,4,5\}$, 5 million pairs are used as training examples, resulting in a total training dataset of 20 million function pairs, which is 4 times higher than used in \citep{anthes2025transformer}, where TSGP was trained for SR problems of a specific dimensionality ($d=4$).

The transformer follows the default architecture and settings chosen in \citep{vaswani2017attention}, with only minor modifications. Because, unlike natural-language tasks, symbolic regression has a considerably smaller vocabulary, we set the embedding dimension to 128 and employ only 2 encoder-decoder stacks, resulting in a model with 3.8M parameters. Preliminary experiments indicated that a larger model capacity does not improve performance on this task. The vocabulary corresponds to the search configuration in Table ~\ref{tab:gp_parameters}, including the primitive set, variables up to dimensionality 5, ERCs and additional integers from \num{-5} to \num{5} (that emerged during SymPy's expression normalization). The full vocabulary is provided in Appendix~\ref{appendix:Vocabulary} (Table~\ref{tab:vocabulary}). Model training was conducted on a NVIDIA TITAN RTX GPU using AdamW optimizer \citep{loshchilov2017decoupled} with the default learning rate of \(10^{-3}\) and a cosine scheduler over 8 epochs \footnote{Pre-trained model and training data are available online: \url{https://doi.org/10.6084/m9.figshare.30601235}}. This training process required approximately 26 hours, but ablation studies showed that similar performance can be achieved more efficiently (compare Appendix \ref{par:Impact_Training_Settings}).

After training, the same trained model is employed in all experiments. During variation, TSGP processes the entire population in a single batch, generating 100 offspring programs in 5-10 seconds (depending on program size) on identical hardware. We evaluate TSGP under different settings of the target semantic distance $\mathrm{SD}_t$. For clarity, we denote the variant with $\mathrm{SD}_t = 1$ as TSGP1, which serves as our default configuration. Other variants are labeled accordingly (e.g. TSGP0.1 for $\mathrm{SD}_t = 0.1$, TSGP5 for $\mathrm{SD}_t = 5$). Unless otherwise specified, “TSGP” in the results refers to TSGP1. Note that during the search, no SymPy is performed. 

\paragraph{Baseline Configurations.}
StdGP uses subtree crossover with a probability of 90\% and a bias toward selecting terminals of 10\%, along with subtree mutation applied with a probability of 10\%, where newly generated subtrees have a depth ranging from 0 to 2 \citep{koza1992programming}. 

For SLIM\_GSGP, we use the default configuration of the framework. Thus, the SLIM+SIG2 mutation operator (which uses addition of two random substructures to create the mutation effect) is used, and no geometric crossover is performed, resulting in a one-parent-based semantic search that makes it comparable to TSGP. The default inflation mutation rate is set to 0.2, defining the probability of selecting the inflation mutation over the deflate operation during the mutation process of a program.

DSR follows the default configuration of the framework. The reward is defined as the negative RMSE to match the metric to measure solution quality, with a batch size of 100 and trained for 100 iterations to align with the settings used in GP.

DAE-GP is implemented as suggested by \citep{wittenberg_denoising_2023}, using an LSTM-based autoencoder with a single hidden layer whose dimensionality is dynamically adjusted. Training is performed using the Adam optimizer with a learning rate of 0.001 until convergence. Input corruption is applied using Levenshtein tree edit at a high edit percentage of 95\%, which promotes extensive structural perturbations and encourages exploration in the search space, helping to counteract premature convergence \citep{wittenberg_denoising_2023}

\section{Experiments} \label{sec:Experiments}
\subsection{Prediction Quality \& Convergence}
\begin{table}
\centering
\caption{Median test RMSE of the best programs (solutions) identified within 100 generations for TSGP with $\mathrm{SD}_t = 1$ , stdGP, SLIM\_GSGP (SLIM), DSR, and DAE-GP (DAE) for the 24 analyzed datasets. Bold values indicate the best prediction quality (lowest RMSE). Significant differences of the best results are indicated by the label symbols.}
\label{tab:prediction_results}

\begin{tabular}{p{0.3\textwidth}|r|r|r|r|r}
\hline
\rowcolor{gray!20}
\textbf{Dataset} & \textbf{\(_a\text{TSGP1}\)} & \textbf{\(_b\text{stdGP}\)} & \textbf{\(_c\text{SLIM}\)} & \textbf{\(_d\text{DSR}\)} & \textbf{\(_e\text{DAE}\)} \\ 

\hline
Analcatdata Apnea1 & 0.9592 & \textbf{\(_{e}\text{0.8176}\)}& 0.9643 & 1.0705 & 1.0662 \\
\rowcolor{gray!10}
Analcatdata Apnea2 & 0.9277 & \textbf{\(_{acde}\text{0.6785}\)} & 1.0287 & 1.2835 & 1.0234\\
Analcatdata Neavote & \textbf{\(_{c}\text{0.2358}\)} & 0.2599 & 0.3536 & 0.2383 & 0.2505 \\
\rowcolor{gray!10}
Chscase Geyser1 & 0.4988 & 0.4906 & 0.4969 & \textbf{\(_{c}\text{0.4732}\)} & 0.4860 \\
Cloud & 0.4601 & 0.5653 & 0.5180 & \textbf{\(_{bce}\text{0.3557}\)} & 0.6248 \\
\rowcolor{gray!10}
ERA & 0.8006 & 0.8250 & \textbf{\(_{bde}\text{0.7943}\)} & 0.9003 & 0.8979\\
ESL & \textbf{\(_{bde}\text{0.3781}\)} & 0.5192 & 0.4138 & 0.4613 & 0.5665 \\
\rowcolor{gray!10}
Feynman I 18 4 & \textbf{\(_{bcde}\text{0.1679}\)} & 0.3124 & 0.2059 & 0.4354 & 0.5344 \\
Feynman I 24 6 & \textbf{\(_{bcde}\text{0.1617}\)} & 0.3752 & 0.2838 & 0.6824 & 0.7271\\
\rowcolor{gray!10}
Feynman I 25 13 & 0.1893 & 0.2078 & \textbf{\(_{de}\text{0.1891}\)} & 0.4736 & 0.5152\\
Feynman I 29 4 & \textbf{\(_{cde}\text{0.2843}\)} & 0.3261 & 0.3405 & 0.5770 & 0.5922 \\
\rowcolor{gray!10}
Feynman I 43 43 & \textbf{\(_{bcde}\text{0.3703}\)} & 0.5182 & 0.4312 & 0.8031 & 0.8103\\
Feynman II 34 2 & \textbf{\(_{bcde}\text{0.0881}\)} & 0.2253 & 0.2014 & 0.6618 & 0.6854 \\
\rowcolor{gray!10}
Feynman II 38 3 & \textbf{\(_{bcde}\text{0.3141}\)} & 0.5260 & 0.4123 & 0.7659 & 0.8289\\
Feynman II 4 23 & \textbf{\(_{bcde}\text{0.2836}\)} & 0.3926 & 0.3936 & 0.7245 & 0.7713\\
\rowcolor{gray!10}
Feynman III 14 14 & \textbf{\(_{bcde}\text{0.4377}\)} & 0.5384 & 0.5405 & 0.9377 & 0.8753\\
Feynman Test 3 & \textbf{\(_{bcde}\text{0.1776}\)} & 0.2391 & 0.2329 & 0.4765 & 0.5511\\
\rowcolor{gray!10}
Feynman Test 4 & \textbf{\(_{cde}\text{0.2211}\)} & 0.2864 & 0.2490 & 0.4698 & 0.5779\\
Galaxy & \textbf{\(_{cde}\text{0.2824}\)} & 0.3145 & 0.3435 & 0.3874 & 0.4491\\
\rowcolor{gray!10}
LEV & \textbf{\(_{bcde}\text{0.6568}\)} & 0.7166 & 0.6629 & 0.8223 & 0.8265\\
Pollen & \textbf{\(_{bcde}\text{0.4700}\)} & 0.5038 & 0.5139 & 0.7259 & 0.8005 \\
\rowcolor{gray!10}
Rabe & \textbf{\(_{bcde}\text{0.0929}\)} & 0.1427 & 0.2243 & 0.2454 & 0.3005\\
Vinnie & 0.5127 & 0.5154 & 0.5121 & \textbf{\(_{abce}\text{0.4862}\)} & 0.5235 \\
\rowcolor{gray!10}
Visualizing Environmental & 0.8213 & 0.8122 & 0.8391 & \textbf{\(_{abce}\text{0.6871}\)} & 0.8174 \\

\hline
\hline
\rowcolor{gray!20}
\textbf{Rank (Mean \& Std)}& \textbf{\(_{bcde}\text{1.58}\)$\pm$1.04} & 2.71$\pm$0.84 & 2.67$\pm$1.03& 3.54$\pm$1.29 & 4.50 $\pm$0.87\\
\end{tabular}
\end{table}

We start our analysis with a performance comparison of TSGP against the benchmark methods stdGP, SLIM\_GSGP, DSR, and DAE-GP. For TSGP, we set the target semantic distance to $\mathrm{SD}_t = 1$ (denoted as TSGP1). In each run, the best-performing program on the training set after 100 generations is selected, and its generalization performance is evaluated using the RMSE on the held-out test set. Table~\ref{tab:prediction_results} reports the median test RMSE over 30 independent runs for each dataset. The corresponding median training RMSE is reported in Table~\ref{tab:training_results} in Appendix~\ref{appendix:Training_RMSE}. Additionally, we compute the average rank of each algorithm across all datasets, along with the standard deviation. The best results are highlighted in bold font and are tested for statistical significance using the Mann-Whitney U test with $\alpha = 0.05$ and Bonferroni correction for multiple comparisons. The method labels ($a$,$b$,$c$,$d$,$e$) highlight statistically significant differences to the worse-performing methods.

The results show that TSGP achieves the best overall performance, significantly outperforming all baselines. Across all datasets, TSGP achieves the lowest average rank (1.58), with statistically significant improvements over stdGP (rank 2.71), SLIM\_GSGP (rank 2.67), DSR (rank 3.54), and DAE-GP (rank 4.50). 

TSGP’s strong performance is robust across dataset characteristics. TSGP consistently produces the best solutions regardless of dimensionality, data type (synthetic or real-world), or sample size. In contrast, the other model-based methods DSR and DAE-GP perform overall poorly, with average ranks of 3.54 and 4.50, respectively. Both rely on online model training during the evolutionary search, which requires a large amount of training samples within the search to generate promising candidate solutions. TSGP, on the other hand, operates offline using a pre-trained model that leverages prior knowledge, making it significantly more efficient at finding high-quality solutions.

Given the inferior performance of DSR and DAE-GP, we focus the remainder of our analysis on the methods that perform best, TSGP, stdGP, and SLIM\_GSGP.
\begin{figure}
  \centering
  \includegraphics[width=\textwidth]{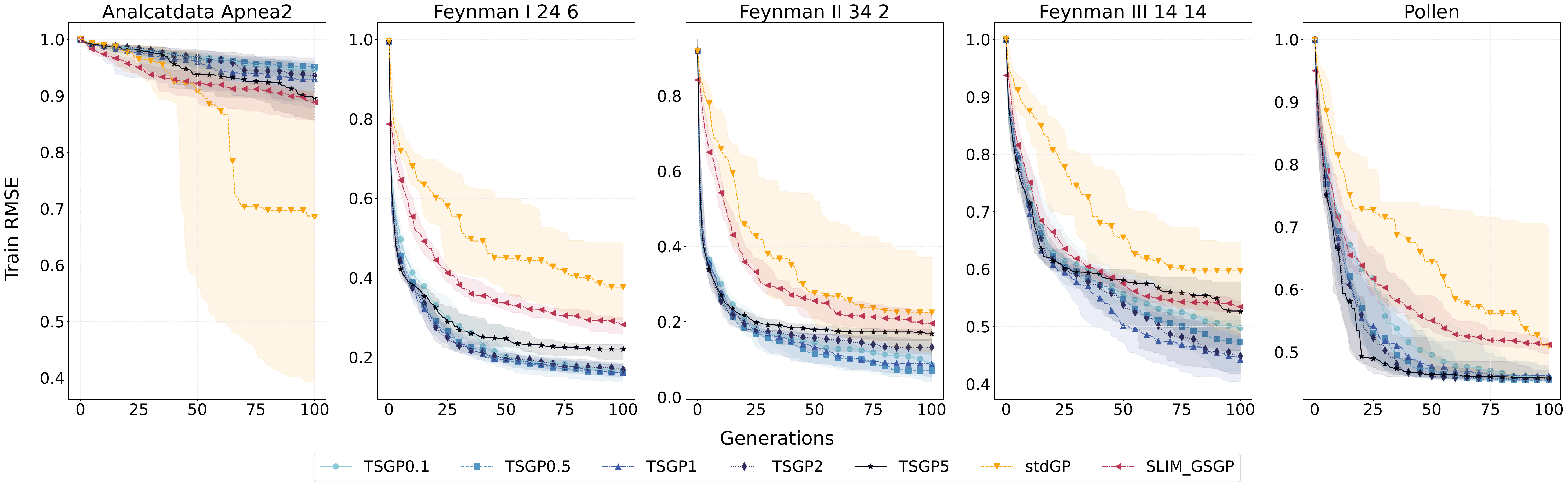} 
  \caption{Median training RMSE of the best programs (solutions) of TSGP, stdGP, SLIM\_GSGP over generations on a subset of the analyzed datasets.}
  \label{fig:Train_Performance}
\end{figure}
We study the convergence speed of TSGP, stdGP, and SLIM\_GSGP to assess how quickly each algorithm discovers programs with low error during the search. This reflects the effectiveness of their variation operators in improving candidate solutions. Consequently, Figure~\ref{fig:Train_Performance} plots the median training RMSE of the best programs over the number of generations. The interquartile range (IQR) indicates performance variability across 30 runs. For clarity, we show five representative datasets; results for the remaining datasets are provided in Appendix~\ref{appendix:Training_RMSE} (Figure~\ref{fig:Train_Performance_All}). In addition to the standard setting (TSGP1), we analyze TSGP under varying $\mathrm{SD}_t$, which influences the degree of semantic similarity between parent and offspring. Small $\mathrm{SD}_t = 0.1$ encourages high semantic similarity, while larger values (up to $\mathrm{SD}_t = 5$) promote greater dissimilarity. These variants are labeled in the legend and visualized using a sequential color scale from light blue (low $\mathrm{SD}_t$) to black (high $\mathrm{SD}_t$).

We start the analysis with the standard variant TSGP1, setting $\mathrm{SD}_t = 1$. TSGP1 converges significantly faster than stdGP and SLIM\_GSGP for almost all datasets and quickly identifies high-quality solutions, which are continuously improved throughout the search. For example, on Feynman I 24 6, TSGP1 finds better solutions than stdGP within just 9 generations and surpasses the termination performance of SLIM\_GSGP by generation 21. In contrast to semantic-based methods, stdGP exhibits slower convergence, reflecting the inefficiency of syntactic variation operators that ignore the semantic effects. 

Furthermore, there are notable differences in convergence behavior even among the semantic methods. TSGP consistently converges significantly faster towards good solutions during the early stages of the search than SLIM\_GSGP. We attribute TSGP’s faster early convergence to its role as a pre-trained semantic operator. During the \textit{Model Building} phase, the transformer model learns useful program structures of the relevant solution space, allowing it to transfer this external knowledge to new problems during the search. In contrast, SLIM\_GSGP operates strictly online and does not benefit from these learned priors.

Notably, the IQR of TSGP and SLIM\_GSGP is smaller than that of stdGP, indicating greater stability and consistency in solution quality across runs. This stability arises because semantic-aware methods explore directly in the semantic solution space and are not based on random structural changes that have uncertain behavioral effects, leading to more consistent and reliable results over multiple runs.

The analysis of TSGP under varying $\mathrm{SD}_t$ values reveals a trade-off between convergence speed and final solution quality. TSGP with small $\mathrm{SD}_t$ often converges more slowly but achieves the highest final performance. In contrast, high $\mathrm{SD}_t$ often leads to faster initial progress (especially on datasets like Pollen and Feynman III 14 14), but frequently results in early plateaus (e.g. on Feynman I 24 6), limiting further improvement. TSGP with $\mathrm{SD}_t = 1$ offers a robust balance, achieving consistently strong convergence behavior and high prediction accuracy across a wide range of problems.

\subsection{Solution Size}

\begin{figure}
  \centering
  \includegraphics[width=\textwidth]{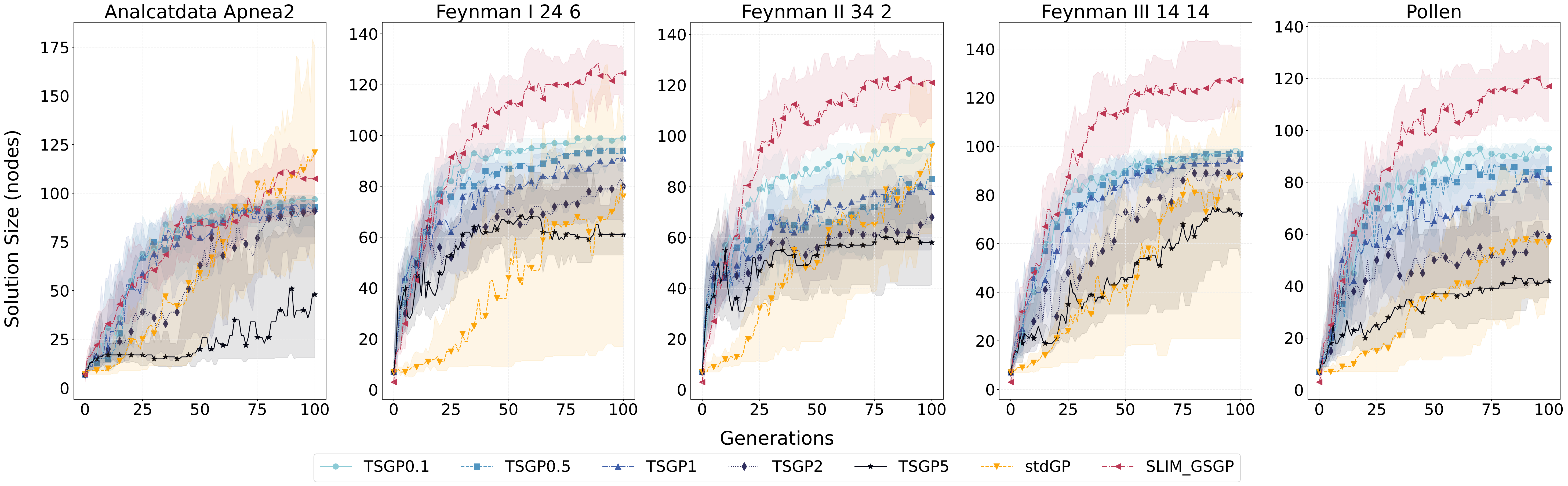} 
  \caption{
 Median size of the solutions of TSGP, stdGP, SLIM\_GSGP over generations on a subset of the analyzed datasets.
 \label{fig:Size}
 }
\end{figure}
Beyond prediction accuracy, program size is a relevant property of solutions, as smaller ones tend to be more interpretable \citep{wittenberg_small_2023,javed2022simplification}. Usually, the size of symbolic expressions is measured by counting the nodes in the expression tree. Table~\ref{tab:solution_size} reports the median size of the solutions after 100 generations. 
First, we compare the solution sizes (sizes of the best programs) generated by the default variant TSGP1 with those produced by stdGP and SLIM\_GSGP. The results are listed in Table~\ref{tab:solution_size}. As before, these values represent the median of 30 independent runs. The best results (smallest solutions) are shown in bold and are tested for significant differences with solution sizes of the other methods. Significant differences are marked with representative labels ($a$,$b$,$c$).

The results show that TSGP1 generates the smallest solutions on average, with a mean rank of 1.58, followed by stdGP (1.75) and SLIM\_GSGP (2.67). This is particularly noteworthy because TSGP1 achieves this compactness while also achieving the best predictive performance.

Furthermore, there are significant differences when comparing the semantic methods. TSGP1, which uses a transformer model to generate semantic similarities, produces significantly smaller solutions than SLIM\_GSGP, which uses linear combination-based GSOs for variation. Being trained on millions of diverse syntactic variations that lead to semantically similar offspring, TSGP discovers solutions that are not only accurate but also more compact than those generated by GSO-based approaches.
\begin{table} 
\centering
\caption{Median program size of the solutions identified within 100 generations by TSGP with $\mathrm{SD}_t = 1$, stdGP, and SLIM\_GSGP (SLIM). Values in bold indicate the smallest solutions. Significant differences with respect to all other methods are indicated by the label symbols.}
\label{tab:solution_size}
\begin{tabular}{p{0.3\textwidth}|r|r|r}\hline
\rowcolor{gray!20}
\textbf{Dataset} & \textbf{\(_a\text{TSGP1}\)} & \textbf{\(_b\text{stdGP}\)} & \textbf{\(_c\text{SLIM}\)} \\ 
 \hline
Analcatdata Apnea1 & \textbf{\(_{c}\text{90}\)} & 94 & 120 \\
\rowcolor{gray!10}
Analcatdata Apnea2 & \textbf{\(_{c}\text{92}\)} & 121 & 108 \\
Analcatdata Neavote & 86 & 105 & \textbf{\(_{}\text{84}\)} \\
\rowcolor{gray!10}
Chscase Geyser1 & 87 & 113 & \textbf{\(_{}\text{71}\)} \\
Cloud & \textbf{\(_{c}\text{74}\)} & 97 & 107 \\
\rowcolor{gray!10}
ERA & 85 & \textbf{\(_{}\text{72}\)} & 90 \\
ESL & 79 & \textbf{\(_{ac}\text{7}\)} & 100 \\
\rowcolor{gray!10}
Feynman I 18 4 & 64 & \textbf{\(_{c}\text{25}\)} & 100 \\
Feynman I 24 6 & 91 & \textbf{\(_{c}\text{76}\)} & 124 \\
\rowcolor{gray!10}
Feynman I 25 13 & \textbf{\(_{c}\text{83}\)} & 108 & 122 \\
Feynman I 29 4 & \textbf{\(_{bc}\text{81}\)} & 101 & 128 \\
\rowcolor{gray!10}
Feynman I 43 43 & 93 & \textbf{\(_{c}\text{57}\)} & 126 \\
Feynman II 34 2 & \textbf{\(_{c}\text{78}\)} & 96 & 121 \\
\rowcolor{gray!10}
Feynman II 38 3 & 90 & \textbf{\(_{}\text{86}\)} & 126 \\
Feynman II 4 23 & \textbf{\(_{c}\text{97}\)} & 108 & 122 \\
\rowcolor{gray!10}
Feynman III 14 14 & 95 & \textbf{\(_{c}\text{88}\)} & 127 \\
Feynman Test 3 & 83 & \textbf{\(_{c}\text{76}\)} & 106 \\
\rowcolor{gray!10}
Feynman Test 4 & \textbf{\(_{c}\text{69}\)} & 73 & 104 \\
Galaxy & \textbf{\(_{c}\text{86}\)} & 92 & 98 \\
\rowcolor{gray!10}
LEV & \textbf{\(_{c}\text{85}\)} & 98 & 109 \\
Pollen & 80 & \textbf{\(_{ac}\text{57}\)} & 117 \\
\rowcolor{gray!10}
Rabe & 83 & \textbf{\(_{}\text{63}\)} & 80 \\
Vinnie & 78 & 99 & \textbf{\(_{a}\text{50}\)} \\
\rowcolor{gray!10}
Visualizing Environmental & \textbf{\(_{c}\text{93}\)} & 99 & 127 \\
\hline
\hline
\rowcolor{gray!20}
\textbf{Rank (Mean \& Std)}& \textbf{\(_{c}\text{1.58}\)}$\pm$0.57 & 1.75$\pm$0.9 & 2.67$\pm$0.69\\
\hline
\end{tabular}
\end{table}

Figure \ref{fig:Size} plots the median size of the best programs over the number of generations. TSGP1 and its variants with varying $\mathrm{SD}_t$ are compared with stdGP and SLIM\_GSGP. The plot displays the median values and IQR across 30 independent runs. For consistency and clarity, the same subset of datasets shown in the previous section is used; the results for the remaining datasets are provided in Appendix~\ref{appendix:Solution_Size} (Figure~\ref{fig:Size_All}).

We first compare TSGP1 with the baseline methods. In early generations, where TSGP1 quickly identifies high-quality solutions, the solution size increases significantly, comparable to the growth observed with SLIM\_GSGP. However, this growth decreases in later generations, even though the quality of the solutions continuously improves. This indicates that the transformer model of TSGP successfully learned a variety of structural semantic modifications, allowing it to create semantically similar offspring of similar size to the parent. SLIM\_GSGP also shows reduced growth in later generations, due to its deflate operator, but the effect is much weaker. This is because SLIM\_GSGP relies on fixed syntactic rules for variation, which limits its ability to adaptively modify the structure of the program while preserving semantics.

For the different TSGP variants, Figure \ref{fig:Size} reveals that the solution size depends on $\mathrm{SD}_t$. TSGP with larger $\mathrm{SD}_t$ produces significantly smaller solution structures than TSGP with small $\mathrm{SD}_t$. This effect is particularly evident on the Pollen dataset, where all variants of TSGP achieve similar final predictive performance at termination, yet their solution sizes differ drastically, ranging from 42 nodes with $\mathrm{SD}_t = 5$ to 93 nodes with $\mathrm{SD}_t = 0.1$. This illustrates that larger semantic steps lead to the discovery of more compact expressions, while small steps lead to higher solution quality at the cost of increased structural complexity. 

\subsection{Step Size in the Semantic Space}
\begin{figure}[t]
 \centering
 \includegraphics[width=\textwidth]{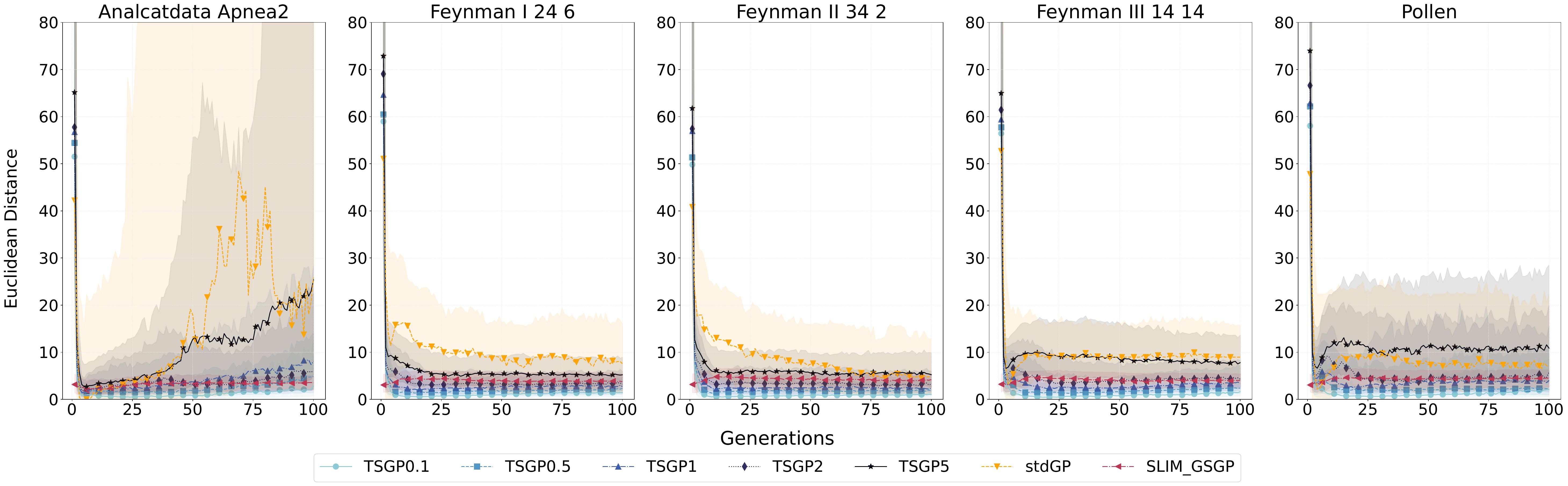} 
 \caption{Median Euclidean distance between the semantics \(s(f_i)\) and \(s(f_o)\) for TSGP with varying $\mathrm{SD}_t$, stdGP, SLIM\_GSGP over the number of generations on a subset of the analyzed datasets.}
 \label{fig:Euclidean_Distance}
\end{figure}
The step size measures the extent of the behavioral changes introduced during variation. It is relevant for the interplay between exploration and exploitation during search. Large step sizes favor exploration, enabling search to explore new areas in the solution space leading to a higher probability of finding the actual global optima \citep{Rothlauf_DesignModernHeuristics_2011}. In contrast, smaller step sizes lead to stronger exploitation (local search) and a gradual improvement in solution quality. Therefore, a successful search strategy must balance both aspects. In TSGP, this balance is regulated by the target semantic distance $\mathrm{SD}_t$, which conditions the transformer model to generate offspring with a specified degree of semantic distance to the parent.

We analyze step size from two perspectives: First, we analyze how $\mathrm{SD}_t$ influences the actual semantic step size, and second, we study how frequently new best solutions are discovered during a run. We focus on the same datasets as before; results for the remaining datasets are provided in Appendix~\ref{appendix:Step_Size}.

We measure the semantic distance $\mathrm{SD}(f_i,f_o)$ between a parent solution \(f_i\) and its offspring \(f_o\) by using the Euclidean distance between their semantic vectors $s(f_i)$ and $s(f_o)$. Small Euclidean distances indicate small steps in semantic space and high semantic similarity; larger distances correspond to larger steps and lower similarity. We calculate the semantic vectors for a fixed, standardized input matrix of randomly sampled points (zero mean, unit variance), analogously to the approach used in \textit{Model Building}. This allows us to study semantic similarities between datasets. Only successful variations are considered in the analysis, where the offspring differs structurally from its parent.

Figure~\ref{fig:Euclidean_Distance} plots the median semantic distance and IQR over the number of generations for TSGP (with varying $\mathrm{SD}_t$), stdGP, and SLIM\_GSGP. TSGP variants are color-coded on a continuous scale from light blue (small $\mathrm{SD}_t$) to black (large $\mathrm{SD}_t$). We show results averaged over 30 runs. A smaller Euclidean distance corresponds to higher semantic similarity. 
\begin{figure}[t]
 \centering
 \includegraphics[width=\textwidth]{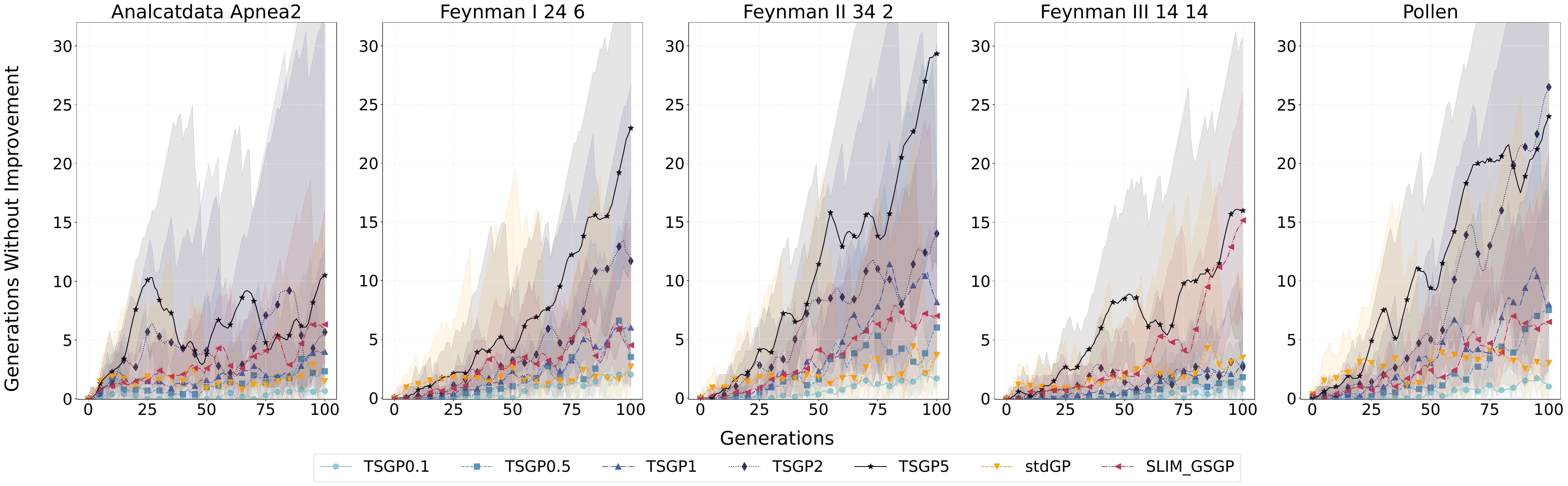} 
 \caption{Median number of generations without improving the training RMSE over the number of generations. Results are for TSGP with varying $\mathrm{SD}_t$, stdGP, and SLIM\_GSGP.}
 \label{fig:Gens_no_improve}
\end{figure}
TSGP1 generates offspring that are semantically much closer to their parental program than those produced by stdGP or SLIM\_GSGP. This is evident from the consistently lower median Euclidean distance across generations, confirming that the transformer has successfully learned to express semantic similarity through syntactic transformations and can generalize this capability to unseen datasets.

An exception is the beginning of the search, where TSGP generates offspring with high semantic dissimilarity. This is due to the random initialization of the initial population, which consists of a variety of programs with a very poor fit to the underlying problem. Since TSGP was primarily trained on training data relevant to the search space, the pre-trained semantic approach cannot find semantic similarities for these random initial programs. 

In contrast, stdGP produces large semantic modifications with high median distances and particularly high 75th percentiles throughout the search. This is an expected behavior, as standard mutation and crossover operators ignore the behavioral change of programs. SLIM\_GSGP maintains a relatively stable semantic similarity with low IQR across generations because its geometric semantic operators explicitly preserve the program behavior. Thus, both semantic methods, TSGP and SLIM\_GSGP, exhibit more stable and controlled variations in program behavior compared to a strict syntactic search like stdGP.

Across all datasets, including those in Figure~\ref{fig:Euclidean_Distance_All} in Appendix~\ref{appendix:Step_Size}, the influence of the target $\mathrm{SD}_t$ on the actual semantic distance is clearly visible. With small $\mathrm{SD}_t$, TSGP produces offspring with low semantic distance (high similarity), while increasing $\mathrm{SD}_t$ leads to larger distances and lower similarity. This shows that $\mathrm{SD}_t$ serves as an effective parameter to influence the semantic step size. Furthermore, in many cases, the actual semantic distance tends to align with the target $\mathrm{SD}_t$, indicating that the transformer has learned to adaptively modulate the semantic similarity between parent and offspring.

In comparison to \citep{anthes2025transformer}, the transformer model has learned to adjust the semantic distance between offspring and parents more precisely. We believe this is due to the influence of SymPy and its syntactic normalization of the training data, which enables a clearer training signal and has a positive effect on the search (as shown in Appendix~\ref{appendix:Ablation_Studies}).

For those datasets where stdGP outperforms the semantic approaches (on Analcatdata Apnea 1 \& 2), stdGP performs particularly large semantic steps during search, which align with its abrupt fitness improvements, as shown in Figure~\ref{fig:Train_Performance}. Notably, these are the only two datasets in which TSGP5 (the variant with the largest target semantic distance) achieves the best performance, while TSGP0.1 (minimal semantic steps) yields the worst results. This suggests that these problems require extensive exploration of the solution space to locate high-quality solutions, providing a scenario in which large, uncontrolled semantic steps (as in stdGP) are advantageous. Instead, the fine-grained and much better regulated step sizes performed with TSGP and SLIM\_GSGP lead to slow convergence. For solving such problems, a stronger randomization of search (higher exploration) using large search steps would be beneficial. 

The setting of $\mathrm{SD}_t$ also strongly affects the frequency with which algorithms discover new candidate solutions during the search. Figure~\ref{fig:Gens_no_improve} plots the median number of generations without finding a new best solution over the number of generations (see Figure~\ref{fig:Gens_no_Improve_All} in Appendix~\ref{appendix:Step_Size} for the remaining datasets). Lower values indicate more frequent improvements. To improve clarity, the curves are smoothed, and the y-axis range is fixed.

We find that TSGP with a small $\mathrm{SD}_t$ (light blue) finds new best solutions in almost every generation, as indicated by the consistently low values. This reflects strong exploitation of the candidate solutions: small semantic adjustments to the parental programs cause a large proportion of the advantageous behavior to be passed on to the offspring. However, these small changes also limit progress, as only minor fitness improvements are possible, resulting in slower convergence (see Figure~\ref{fig:Train_Performance}). Furthermore, it results in larger solution sizes and increased structural complexity (see Figure~\ref{fig:Size}), as high exploitation reduces the ability of TSGP to explore various structural compositions. 
In contrast, larger $\mathrm{SD}_t$ values lead to more exploration. As the semantic distance between parent and offspring is larger, improvements occur less frequently, however, successful search steps often lead to much higher improvements. Especially in the first generations, strong exploration often leads to strong fitness improvements (Figure~\ref{fig:Train_Performance}) and the discovery of more compact solutions (Figure~\ref{fig:Size}). In summary, TSGP with $\mathrm{SD}_t = 1$ achieves a robust balance between exploration and exploitation, leading to strong and consistent performance across diverse problems.

\section{Limitations \& Future Work}\label{sec:Limitations_Future_Work}
While the current implementation is already yielding very promising results, there are still some limitations that hinder its broader applicability and efficiency. 

The current transformer model is constrained to a fixed vocabulary that includes input variables up to dimensionality $d = 5$ and a predefined set of primitives and constants. This limits applicability to problems that require higher-dimensional inputs or different operator sets. Future work will investigate how performance scales with broader vocabularies and higher-dimensional problems through more comprehensive model training.

Additionally, the transformer was pre-trained using programs derived from synthetic datasets with a bias towards smooth, near-linear behaviors. Despite this bias, TSGP performed well on nonlinear and real-world problems, suggesting strong generalization capabilities. Therefore, we plan to investigate the extent to which TSGP can produce semantically similar variants of solutions outside its learned distribution and assess how including more nonlinear behavior in the transformer's training data affects solution quality.

Lastly, semantic methods such as TSGP and SLIM\_GSGP depend on the magnitude of the semantic step size. On certain problems (e.g., Analcatdata Apnea1 \& 2), extensive exploration was required, and low semantic step sizes led to poor performance. Future work will explore adaptive scheduling mechanisms that dynamically adjust the target semantic distance $\mathrm{SD}_t$ during search to better balance exploration and exploitation across diverse problem landscapes.

\section{Conclusions}\label{sec:Conclusions}
This work substantially extends a previous conference paper introducing Transformer Semantic Genetic Programming \citep{anthes2025transformer}, a semantic search method that employs a pre-trained transformer model as a zero-shot variation operator to generate offspring with adjustable semantic similarity to a parent program.

We find that a single transformer model can effectively learn and generalize semantic similarity across SR problems of varying input dimensionality. In a comprehensive evaluation of 24 real-world and synthetic datasets, TSGP achieves an average rank of 1.58, significantly outperforming all other baseline methods, including standard GP (rank 2.71), SLIM\_GSGP (rank 2.67), Deep Symbolic Regression (rank 3.54), and DAE-GP (rank 4.50). Notably, despite its superior performance, TSGP produces solutions of similar compactness to standard GP (average rank 1.58 vs. 1.75) and significantly more compact than those of SLIM\_GSGP (rank 2.67), the bloat-mitigating variant of GSGP.

Further analysis confirms that conditioning the transformer model on a target semantic distance $\mathrm{SD}_t$ systematically influences the actual semantic similarity between parent and offspring. Larger $\mathrm{SD}_t$ values enhance exploration, leading to faster initial convergence and smaller solution structures. Smaller $\mathrm{SD}_t$ values promote exploitation, enabling gradual fitness improvements but often at the cost of increased program size and slower convergence. When using fixed semantic step sizes, TSGP with $\mathrm{SD}_t = 1$ leads to a robust balance between exploitation and exploration and high performance across diverse problems.

\begin{acks}
We would like to thank David Wittenberg for the valuable insights into DAE-GP and for providing his framework. We also like to thank the entire team in Mainz for the inspiring discussions and thoughtful contributions.
\end{acks}
\bibliographystyle{ACM-Reference-Format}
\bibliography{TSGP}
\clearpage
\appendix

\section{TSGP Vocabulary}\label{appendix:Vocabulary}
The vocabulary of the transformer model consists of four main categories: special tokens, arithmetic operations, input variables, and constant values. Special tokens include the empty string, [UNK] (unknown token), [start], and [end], which serve as standard sequence delimiters commonly used in transformer architectures. The set of arithmetic operations consists of standard binary functions: add (addition), sub (subtraction), mul (multiplication), div (division), and pow (exponentiation). Input variables are represented as $x1$ through $x5$, allowing up to five distinct features, though the actual variables used are dynamically adjusted by the model hyperparameter $d$, which matches the problem's dimensionality. Constant values include fractional values from $[-0.5,0.5]$ in steps of 0.1, matching the Ephemeral Random Constant (ERC) range used during the evolutionary search and integers in the ranges $[-5, 5]$ in steps of 1, to maintain consistency with the training data of TSGP, where such values emerge from the normalization of symbolic expressions using SymPy.

\begin{table}[H]
\centering
\renewcommand{\arraystretch}{1.2}
\caption{The vocabulary of the TSGP transformer model consists 
of four main categories: special tokens, arithmetic operations, 
input variables, and constant values.}
\label{tab:vocabulary}
\begin{tabular}{>{\ttfamily}l <{\normalfont} l}
\toprule
\multicolumn{2}{l}{\textbf{Special Tokens}} \\
\midrule
""        & Empty string  \\
{[UNK]}   & Unknown token \\
{[start]}   & Start token \\
{[end]}   & End token \\
\addlinespace

\multicolumn{2}{l}{\textbf{Arithmetic Operations}} \\
\midrule
add       & Addition ($a + b$) \\
sub       & Subtraction ($a - b$) \\
mul       & Multiplication ($a \times b$) \\
div       & Division ($a / b$) \\
pow       & Power ($a^b$) \\
\addlinespace

\multicolumn{2}{l}{\textbf{Input Variables}} \\
\midrule
x1, x2, x3, x4, x5 & Available input features \\
\addlinespace

\multicolumn{2}{l}{\textbf{Constant Values}} \\
\midrule
$[-0.5, 0.5] $ & Floating-point constants (step size 0.1) \\
$[-5, 5]$ & Integers (step size 1)\\
\bottomrule
\end{tabular}
\end{table}

\section{Extended Experimental Results}\label{appendix:Extended_Experimental_Results}
\subsection{Training RMSE}\label{appendix:Training_RMSE}
\begin{table}[H] 
\caption{Median training RMSE of the best programs (solutions) found within 100 generations for TSGP with $\mathrm{SD}_t = 1$ , stdGP, SLIM\_GSGP (SLIM), DSR, and DAE-GP (DAE) for the 24 analyzed datasets. Bold values indicate the best prediction quality (lowest RMSE). Significant differences of the best results are indicated by the label symbols.}
\centering
\label{tab:training_results}
\begin{tabular}{p{0.3\textwidth}|r|r|r|r|r}
\hline
\rowcolor{gray!20}
\textbf{Dataset} & \textbf{\(_a\text{TSGP1}\)} & \textbf{\(_b\text{stdGP}\)} & \textbf{\(_c\text{SLIM}\)} & \textbf{\(_d\text{DSR}\)} & \textbf{\(_e\text{DAE}\)} \\ 

\hline
Analcatdata Apnea1 & 0.9359 & \textbf{\(_{d}\text{0.6768}\)} & 0.8788 & 0.9316 & 0.9915 \\
\rowcolor{gray!10}
Analcatdata Apnea2 & 0.9298 & \textbf{\(_{abce}\text{0.6848}\)} & 0.8886 & 0.9416 & 0.9915 \\
Analcatdata Neavote & \textbf{\(_{bde}\text{0.2122}\)} & 0.2277 & 0.2155 & 0.2512 & 0.2561 \\
\rowcolor{gray!10}
Chscase Geyser1 & \textbf{\(_{de}\text{0.4530}\)} & 0.4657 & 0.4544 & 0.4888 & 0.4859 \\
Cloud & \textbf{\(_{bcde}\text{0.2835}\)} & 0.4099 & 0.3431 & 0.4649 & 0.5166 \\
\rowcolor{gray!10}
ERA & \textbf{\(_{bcde}\text{0.7775}\)} & 0.8077 & 0.7856 & 0.8349 & 0.8901 \\
ESL & \textbf{\(_{bcde}\text{0.3567}\)} & 0.5324 & 0.3846 & 0.4694 & 0.5720 \\
\rowcolor{gray!10}
Feynman I 18 4 & \textbf{\(_{bcde}\text{0.1630}\)} & 0.3145 & 0.2024 & 0.4371 & 0.5300 \\
Feynman I 24 6 & \textbf{\(_{bcde}\text{0.1616}\)} & 0.3768 & 0.2825 & 0.6633 & 0.7275 \\
\rowcolor{gray!10}
Feynman I 25 13 & 0.1934 & 0.2132 & \textbf{\(_{de}\text{0.1817}\)} & 0.4831 & 0.5175 \\
Feynman I 29 4 & \textbf{\(_{cde}\text{0.2815}\)} & 0.3224 & 0.3383 & 0.5796 & 0.5845 \\
\rowcolor{gray!10}
Feynman I 43 43 & \textbf{\(_{bcde}\text{0.3809}\)} & 0.5353 & 0.4260 & 0.7755 & 0.8029 \\
Feynman II 34 2 & \textbf{\(_{bcde}\text{0.0876}\)} & 0.2246 & 0.1957 & 0.6490 & 0.6767 \\
\rowcolor{gray!10}
Feynman II 38 3 & \textbf{\(_{bcde}\text{0.3154}\)} & 0.5196 & 0.4030 & 0.7690 & 0.8307 \\
Feynman II 4 23 & \textbf{\(_{bcde}\text{0.2879}\)} & 0.3978 & 0.3880 & 0.6999 & 0.7417 \\
\rowcolor{gray!10}
Feynman III 14 14 & \textbf{\(_{bcde}\text{0.4421}\)} & 0.5971 & 0.5346 & 0.8361 & 0.8752 \\
Feynman Test 3 & \textbf{\(_{bcde}\text{0.1718}\)} & 0.2328 & 0.2301 & 0.4769 & 0.5491 \\
\rowcolor{gray!10}
Feynman Test 4 & \textbf{\(_{cde}\text{0.2237}\)} & 0.2797 & 0.2479 & 0.4717 & 0.5803 \\
Galaxy & 0.2739 & 0.3055 & \textbf{\(_{de}\text{0.2641}\)} & 0.3979 & 0.4580 \\
\rowcolor{gray!10}
LEV & \textbf{\(_{bcde}\text{0.6536}\)} & 0.7196 & 0.6669 & 0.7564 & 0.8222 \\
Pollen & \textbf{\(_{cde}\text{0.4624}\)} & 0.5102 & 0.5121 & 0.7170 & 0.7836 \\
\rowcolor{gray!10}
Rabe & \textbf{\(_{bde}\text{0.0934}\)} & 0.1814 & 0.0993 & 0.1891 & 0.3448 \\
Vinnie & \textbf{\(_{bde}\text{0.4889}\)} & 0.4939 & 0.5018 & 0.4977 & 0.5042 \\
\rowcolor{gray!10}
Visualizing Environmental & \textbf{\(_{de}\text{0.7364}\)} & 0.7641 & 0.7423 & 0.8241 & 0.8193 \\
\hline
\hline
\rowcolor{gray!20}
\textbf{Rank (Mean \& Std)}& \textbf{\(_{bcde}\text{1.29}\)$\pm$0.73}& 2.75$\pm$0.66& 2.08$\pm$0.57& 3.96$\pm$0.45& 4.92$\pm$0.28\\
\end{tabular}
\end{table}
\clearpage

\begin{figure}
  \centering
  \includegraphics[width=\textwidth]{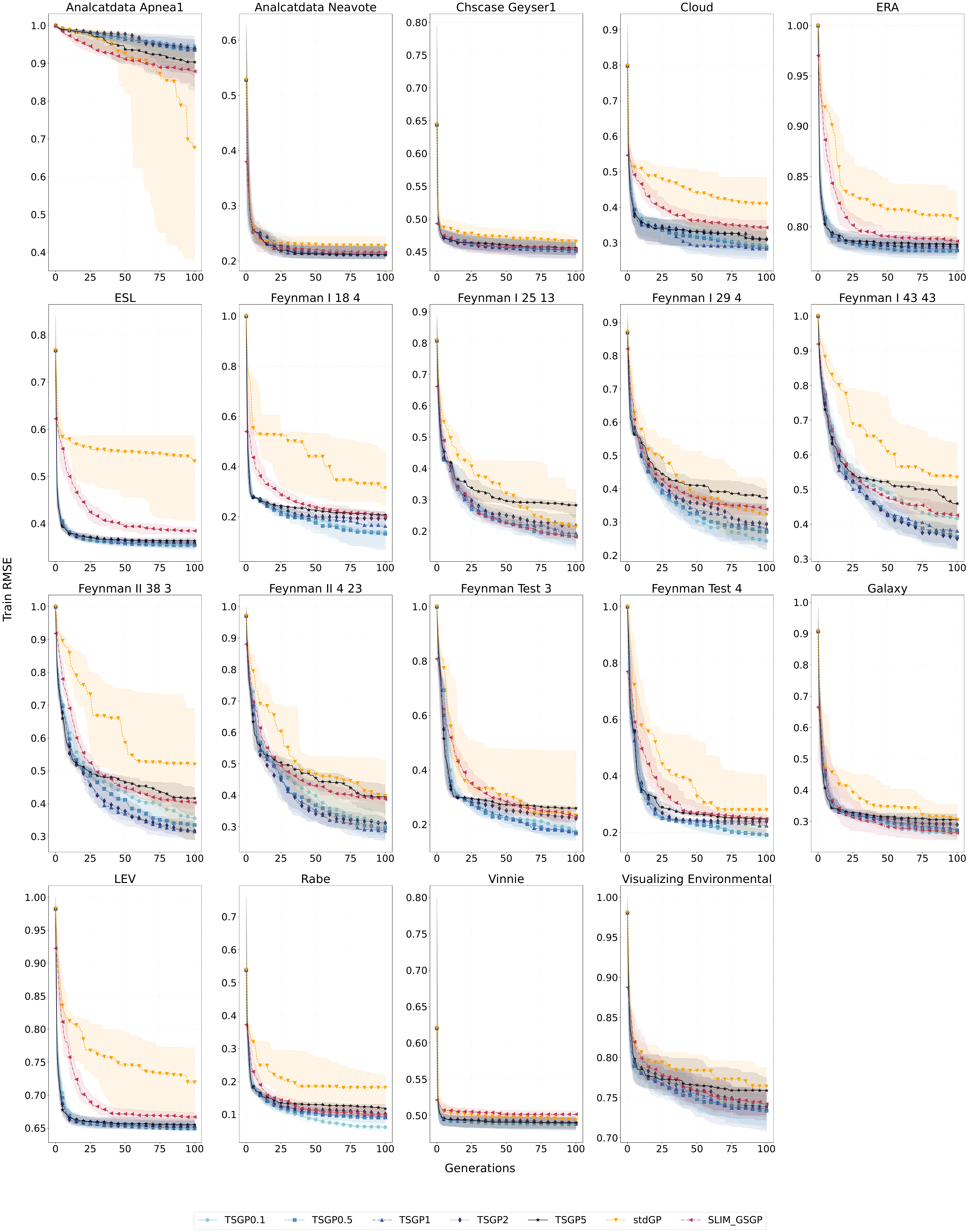} 
 \caption{Median training RMSE of the solutions of TSGP, stdGP, SLIM\_GSGP over the number of generations on the remaining analyzed datasets}
\label{fig:Train_Performance_All}
\end{figure}
\clearpage

\subsection{Solution Size}\label{appendix:Solution_Size}

\begin{figure}[H]
  \centering
  \includegraphics[width=\textwidth]{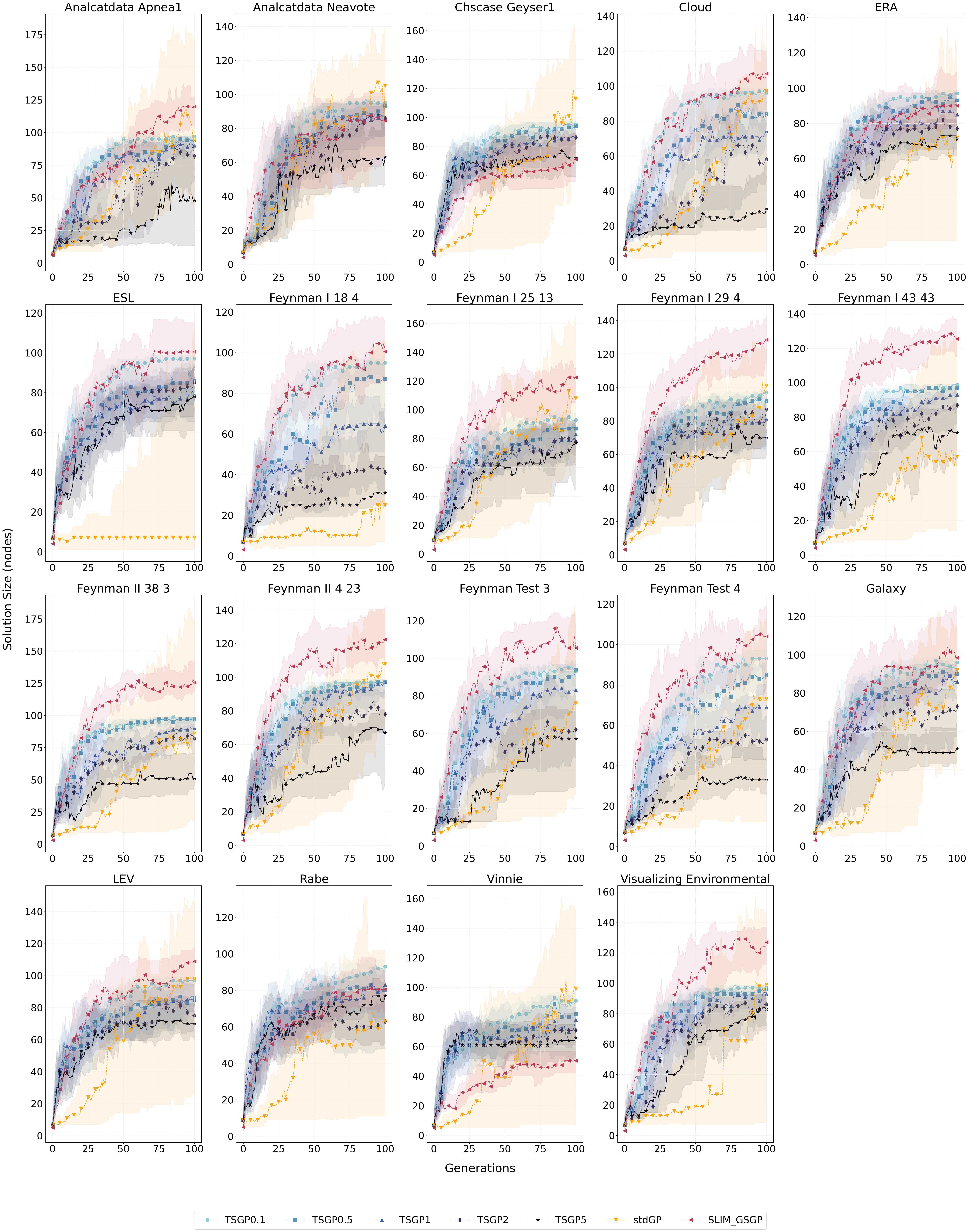} 
   \caption{Median solution size of TSGP, stdGP, SLIM\_GSGP over the number of generations on the remaining analyzed datasets.}
  \label{fig:Size_All}
\end{figure}

\subsection{Step Size}\label{appendix:Step_Size}

\begin{figure}[H]
  \centering
  \includegraphics[width=\textwidth]{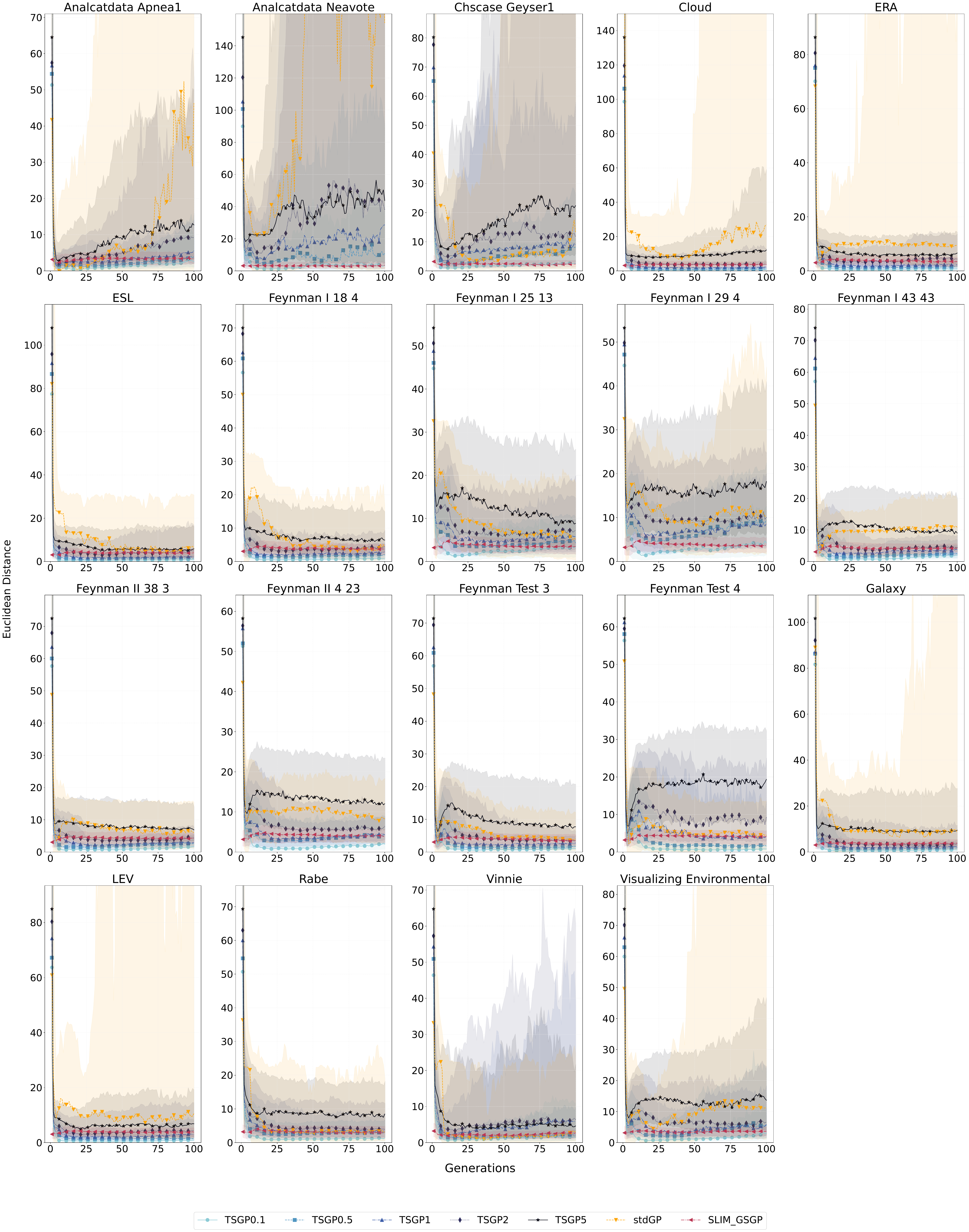} 
  \caption{Median Euclidean distance between the semantics \(s(f_i)\) and \(s(f_o)\) for TSGP with varying $\mathrm{SD}_t$, stdGP, SLIM\_GSGP over the number of generations on the remaining analyzed datasets.}  
  \label{fig:Euclidean_Distance_All}
\end{figure}

\begin{figure}
  \centering
  \includegraphics[width=\textwidth]{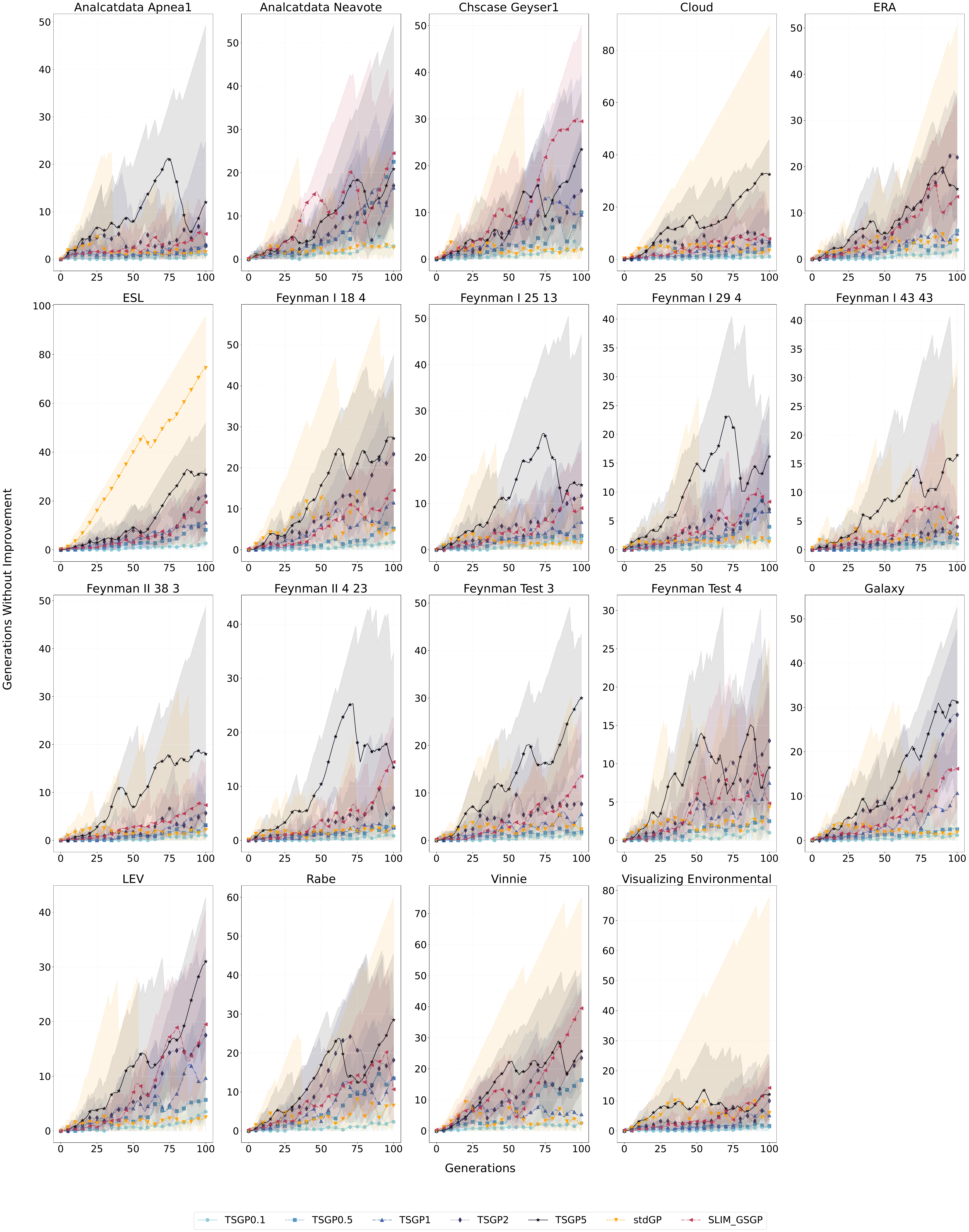} 
  \caption{Median number of generations without improving the training RMSE over the number of generations. Results are for TSGP with varying $\mathrm{SD}_t$, stdGP, and SLIM\_GSGP.}
  \label{fig:Gens_no_Improve_All}
\end{figure}

\section{Ablation Studies}\label{appendix:Ablation_Studies}
We evaluate how different configurations during \textit{Model Building} and \textit{Model Inference} affect the performance of TSGP. Results comparing the baseline TSGP model against all tested variants (with $\mathrm{SD}_t = 1$) on the five representative datasets used in the main study are presented in Table~\ref{tab:ablation_studies}. For changes affecting \textit{Model Building} (and thus the transformer’s training phase), we additionally report training performance in Table~\ref{tab:model_training_results}.
 
\begin{table}[H] 
\caption{Validation performance of the transformer model for different \textit{Model Building} settings. Higher accuracy and lower loss indicate better model performance.}
\centering
\label{tab:model_training_results}
\begin{tabular}{p{0.3\textwidth}|r|r}
\hline
\rowcolor{gray!20}
\textbf{Model Setting} & \textbf{Validation Accuracy} & \textbf{Validation Loss} \\
TSGP & 94.92\% & 0.1585 \\
\rowcolor{gray!10}
TSGP k=100 & 93.62\% & 0.1995 \\
TSGP no SymPy & 92.96\% & 0.2361 \\
\rowcolor{gray!10}

TSGP 10\% Training Data (10\% TD)&  94.32\%& 0.1786\\
TSGP 1 Epoch (1E)& 93.99\% & 0.1925 \\
\rowcolor{gray!10}
TSGP 10\% TD \& 1E& 92.25\%& 0.2479\\

\hline
\end{tabular}
\end{table}
\begin{table}[H] 
\caption{Median test RMSE of the best programs (solutions) found within 100 generations for TSGP with $\mathrm{SD}_t = 1$ and its variants. Results include variations in data quality (k=100, No SymPy), training quantity (10\% data, 1 epoch) and inference settings (No SC, LLM)}
\centering
\label{tab:ablation_studies}
\begin{tabular}{l|r|r|r|r|r|r|r|l}
\hline
\rowcolor{gray!20}
\textbf{Dataset} & \textbf{TSGP} & \textbf{k=100}& \textbf{No SymPy} & \textbf{10\%}& \textbf{1E}& \textbf{10\% \& 1E}&\textbf{No SC} & \textbf{LLM}\\
A. Apnea2 & 0.9277 &1.0209& 0.9902 & 0.8647& 0.9703& 0.9074 & 0.9564&1.2091\\
\rowcolor{gray!10}
Fey. I 24 6 & 0.1617 &0.2073& 0.3103 & 0.1494& 0.2213& 0.1783 &0.1581 &0.7225\\
Fey. II 34 2 & 0.0881 &0.1675& 0.2070 & 0.0965& 0.1295& 0.1289 &0.1123 &0.6559\\
\rowcolor{gray!10}
Fey. III 14 14 & 0.4377 &0.5520& 0.5685 & 0.4588& 0.4682& 0.4718 &0.4688 &0.9290\\
Pollen & 0.4700 &0.4762& 0.6657 & 0.4737& 0.4713& 0.4708 &0.4673 &0.8510\\
\hline

\end{tabular}
\end{table}

\begin{figure} 
    \centering
    \includegraphics[width=1\linewidth]{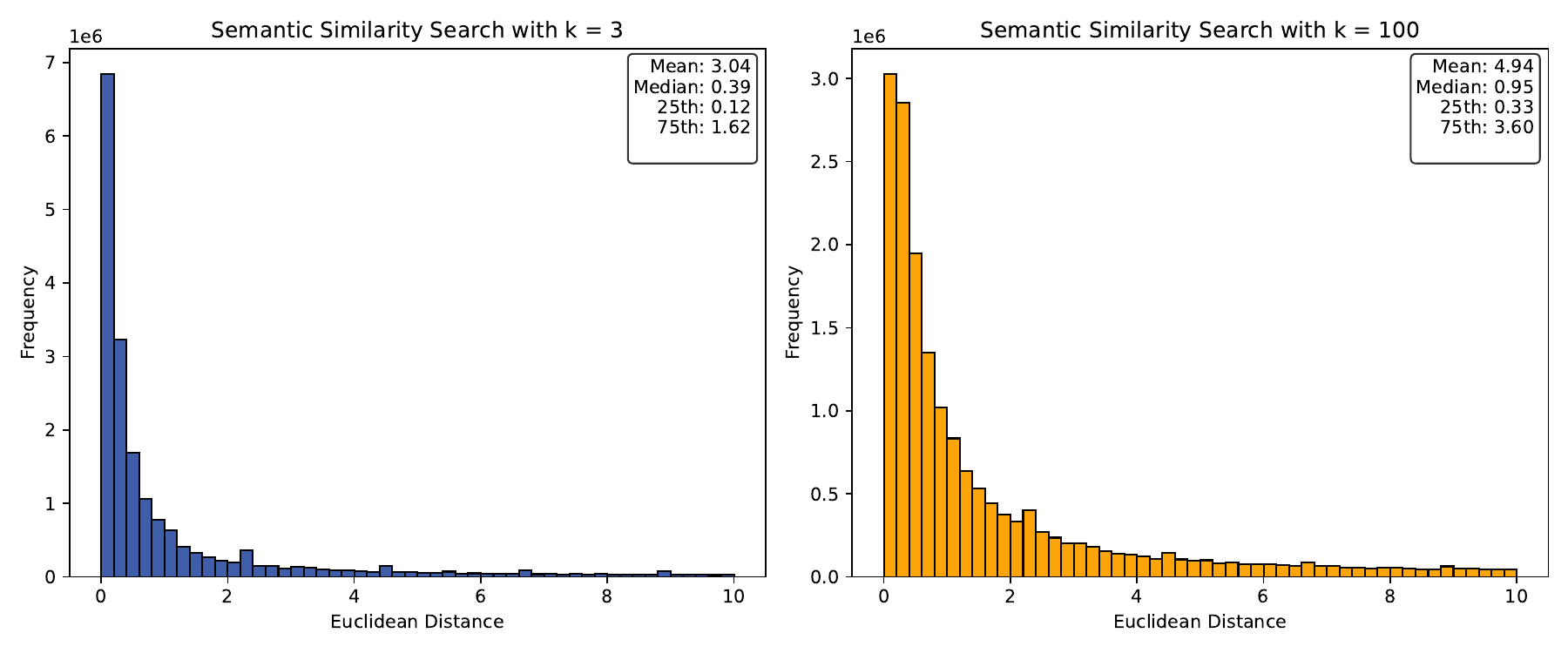}
    \caption{Distribution of semantic distances (Euclidean) in training data for k=3 (baseline) and $k =100$ (ablation). Higher $k$ leads to higher semantic distances in the training data. To illustrate the majority of the distribution, only semantic distances $\mathrm{SD} < 10$ are shown.}
    \label{fig:Distances_Trainingdata}
\end{figure}

\paragraph{Impact of Data Quality.} \label{par:Impact_Data_Quality}
During \textit{Model Building}, training pairs are formed by selecting the $k$ nearest neighbors of each program in the semantic space. Larger $k$ increases the average semantic distance between pairs, providing a weaker signal about what constitutes semantic similarity. To isolate this effect, we created training data using $k$ = 100, but randomly sampled only 3 out of the 100 neighbors per function to match the baseline’s training data size. The resulting distributions of semantic distances for both $k$=3 (baseline) and $k$=100 (ablation) are visualized in Figure~\ref{fig:Distances_Trainingdata}. 

Training on data with higher semantic distances negatively impacts both model training and evolutionary search performance (TSGP k=100). Specifically, validation accuracy drops by 1.3\%, and validation loss increases by 0.041. Importantly, the model consistently underperforms the TSGP baseline across all five datasets during evolutionary search.

Data quality is further influenced by syntactic redundancy (TSGP no SymPy). Normalizing expressions using SymPy eliminates such redundancies and yields a cleaner, more consistent mapping between syntactic and semantic changes. As a result, training performance improves significantly from 92.96\% (without SymPy) to 94.92\% (baseline), and validation loss decreases from 0.2361 to 0.1585. Consequently, the model trained on unnormalized functions performs worse during search, consistently underperforming the baseline TSGP.\footnote{This does not imply equivalence to the model used in \citep{anthes2025transformer}, which also omitted SymPy. In \citep{anthes2025transformer}, the similarity search was performed separately for each archive generated from GP runs of a specific dimensionality (not combined), and the target semantic distance was set to $\mathrm{SD}_t = 0.1$.}

\paragraph{Impact of Training Settings.} \label{par:Impact_Training_Settings}
Next, we examine how model performance is affected by the amount of training data and the duration of training.
To assess data efficiency, we trained a variant using only 10\% of the original training data (2 million function pairs), denoted as "TSGP 10\% Training Data". This reduction only marginally impacted training performance, as reflected in the validation metrics, and did not result in a significant decline of search performance across the evaluated datasets.

The impact of training duration was examined through a single-epoch training variant (TSGP 1 Epoch), with a learning rate scheduler adjusted to maximize convergence within the shortened time-frame. Compared to the data-reduced variant, this setting had a slightly stronger negative impact on both training convergence and search performance. Nevertheless, the effect remained relatively mild when compared to changes affecting data quality. The combined effect of reduced data and shortened training (10\% TD \& 1E) further degraded training metrics but did not produce worse search performance than the individual reductions.

The results demonstrate that the TSGP model is robust to reductions in training data volume or training duration. While search performance does degrade slightly under these conditions, the impact is far less severe than that caused by poor data quality.

\paragraph{Impact of Inference Settings.} \label{par:Impact_Inference_Settings}
During inference, TSGP employs syntax control (SC) to ensure syntactically valid program generation. To evaluate its importance, we tested a variant with SC disabled (“No SC”), where invalid offspring are replaced with the parent program. Disabling SC had variable but generally negative effects, particularly on Feynman II 34 2 and Feynman III 14 14. On other datasets, performance remained comparable. 

Finally, we evaluated whether a general-purpose large language model (LLM) could be a proper substitute for TSGP's specialized transformer. Using Qwen3-4B (approximately 1,000× larger than TSGP's 3.8M-parameter model), we provided a system prompt instructing it to function as a semantic variation operator followed by TSGP's vocabulary in Appendix~\ref{appendix:Vocabulary}: 
"You are a semantic variation operator for genetic programming. Your task: Generate a new mathematical expression that is semantically similar to the input expression, but not equivalent. Only output the expression in prefix notation - nothing else. Use ONLY prefix notation with these tokens: [...]". 

Due to computational constraints, “thinking mode” was disabled, and the maximum token limit was set to 100 (matching TSGP’s transformer). Additionally, we conducted only a limited number of runs (n=5), as sequential sampling of a population is computationally expensive (\~90 seconds on the same hardware). In case of an invalid offspring, the parent function is used as the offspring instead. Preliminary experiments revealed that the LLM proposed only a small number of novel solutions during search, so we tested multiple temperatures ($t$) in a preliminary study and selected the best value ($t = 2$) for all subsequent experiments. Despite its significantly larger parameter count and high temperature, the LLM variant consistently converged prematurely, producing low-quality solutions that performed worse than TSGP across all benchmarks.

\end{document}